\documentclass[journal]{IEEEtran}
\usepackage{times}
\usepackage{epsfig}
\usepackage{graphicx}
\usepackage{amsmath}
\usepackage{amssymb}
\usepackage{animate}
\usepackage{algorithm}
\usepackage{algpseudocode}
\algrenewcommand\algorithmicindent{.9em}%
\usepackage{enumerate}
\usepackage{multirow}
\usepackage{enumitem}
\usepackage{makecell}
\usepackage{tabulary}
\usepackage{pifont}
\usepackage{boldline}
\usepackage{bm}
\usepackage{gensymb}
\usepackage[dvipsnames]{xcolor}
\usepackage{caption}
\usepackage{subcaption}
\usepackage{footmisc}
\usepackage{colortbl}
\usepackage{float}
\usepackage{url}
\usepackage{array}
\usepackage{cite}
\usepackage[colorlinks,urlcolor=black,linkcolor=black,anchorcolor=black,citecolor=black]{hyperref}
\begin{document}
	\title{3D Cascade RCNN:\\ High Quality Object Detection in Point Clouds}
	\author{Qi Cai, Yingwei Pan,~\IEEEmembership{Member,~IEEE,} Ting Yao,~\IEEEmembership{Senior Member,~IEEE,} and Tao Mei,~\IEEEmembership{Fellow,~IEEE}
		\thanks{Qi Cai is with University of Science and Technology of China, Hefei 230026, China (e-mail: caiqicq@mail.ustc.edu.cn).}
		\thanks{Yingwei Pan, Ting Yao and Tao Mei are with JD AI Research, Beijing 100105, China (e-mail:panyw.ustc@gmail.com; tingyao.ustc@gmail.com; tmei@jd.com).}
        \thanks{Corresponding author: Ting Yao.}}
	\maketitle
	\begin{abstract}
Recent progress on 2D object detection has featured Cascade RCNN, which capitalizes on a sequence of cascade detectors to progressively improve proposal quality, towards high-quality object detection. However, there has not been evidence in support of building such cascade structures for 3D object detection, a challenging detection scenario with highly sparse LiDAR point clouds. In this work, we present a simple yet effective cascade architecture, named 3D Cascade RCNN, that allocates multiple detectors based on the voxelized point clouds in a cascade paradigm, pursuing higher quality 3D object detector progressively. Furthermore, we quantitatively define the sparsity level of the points within 3D bounding box of each object as the point completeness score, which is exploited as the task weight for each proposal to guide the learning of each stage detector. The spirit behind is to assign higher weights for high-quality proposals with relatively complete point distribution, while down-weight the proposals with extremely sparse points that often incur noise during training. This design of completeness-aware re-weighting elegantly upgrades the cascade paradigm to be better applicable for the sparse input data, without increasing any FLOP budgets. Through extensive experiments on both the KITTI dataset and Waymo Open Dataset, we validate the superiority of our proposed 3D Cascade RCNN, when comparing to state-of-the-art 3D object detection techniques. The source code is publicly available at \url{https://github.com/caiqi/Cascasde-3D}.
	\end{abstract}
	
	
	\IEEEpeerreviewmaketitle
	
	\section{Introduction}
	
\begin{figure}[!tb]
		\begin{center}
			\begin{subfigure}[t]{0.47\linewidth}
				\centering\includegraphics[width=1.0\linewidth]{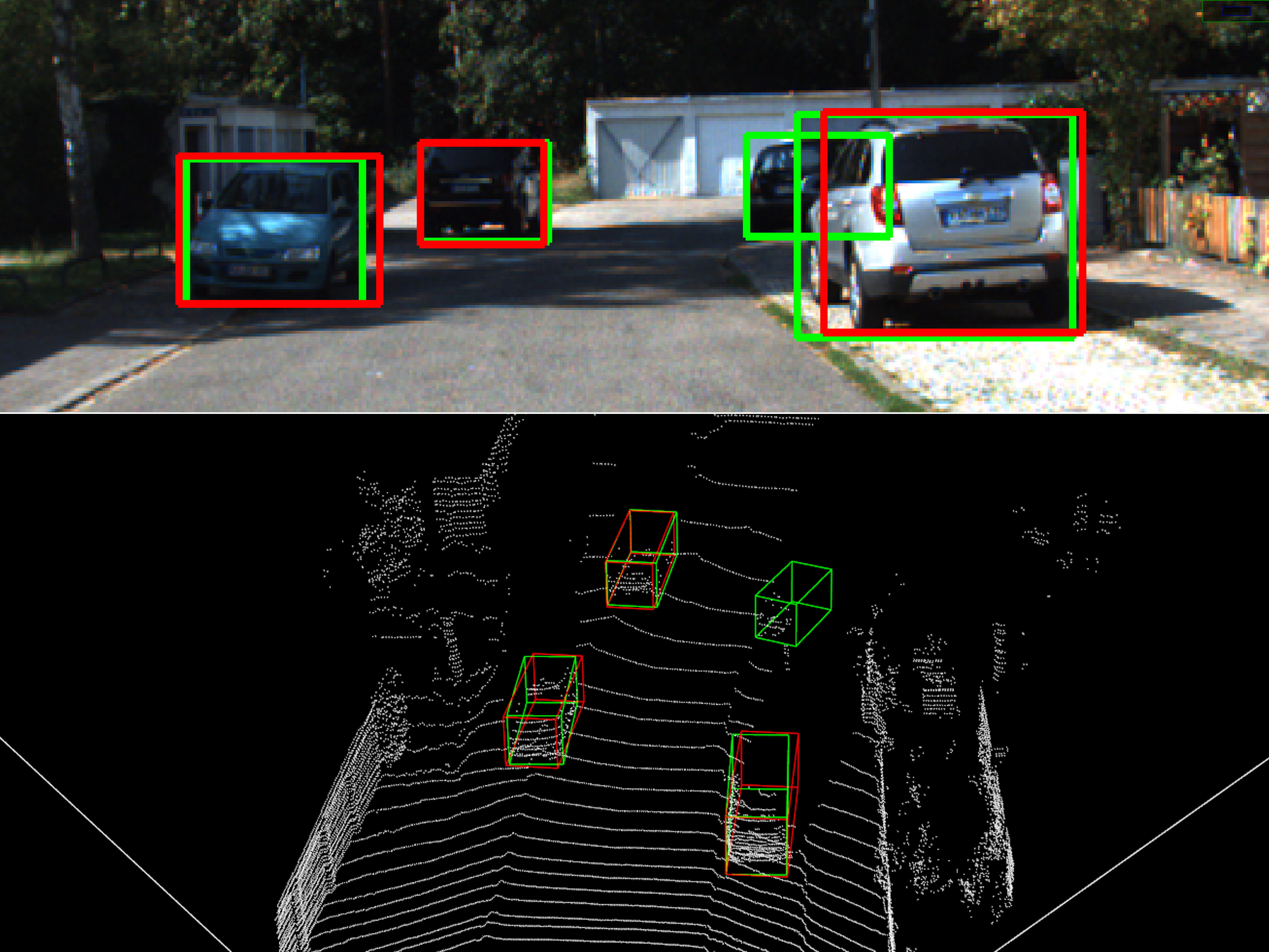}
				\caption{PV-RCNN \cite{shi2020pv}}
			\end{subfigure}
			\begin{subfigure}[t]{0.47\linewidth}
				\centering\includegraphics[width=1.0\linewidth]{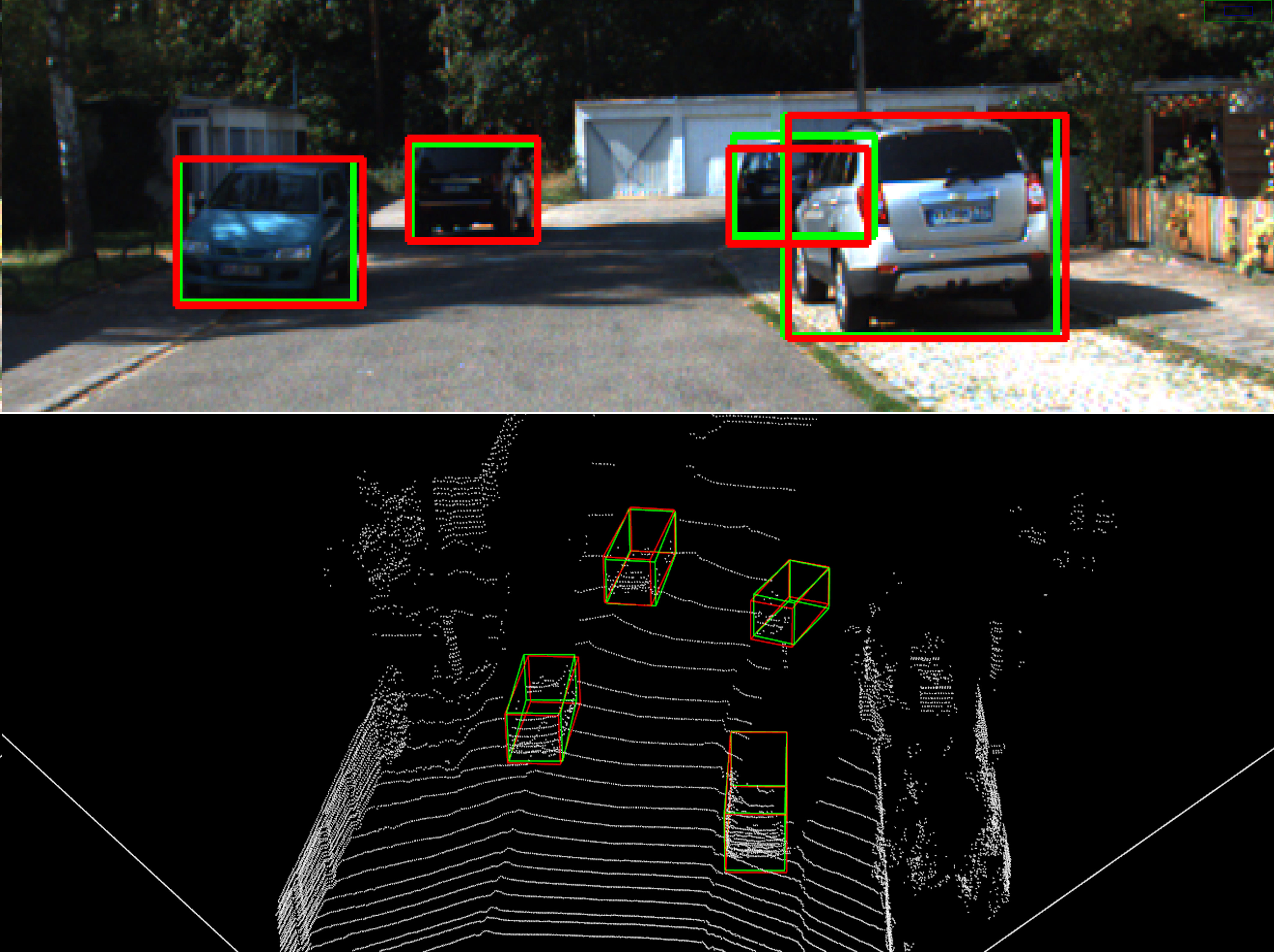}
				\caption{3D Cascade RCNN}
			\end{subfigure}
		\end{center}
		\caption{The upper part is the RGB image and the bottom part is the view of entire point clouds, coupled with the 3D bounding boxes of target objects (\textcolor{green}{GREEN} box: ground truth; \textcolor{red}{RED} box: detection results of PV-RCNN and 3D Cascade RCNN). The PV-RCNN fails to detect the remote car with fewer/sparse points while 3D Cascade RCNN detects all cars.}
		\label{fig:intro}
	\end{figure}

	\IEEEPARstart{3}{D} object detection, \emph{i.e.}, the task of localizing and classifying objects in point clouds, is one of the most fundamental challenges in the computer vision field. Practical 3D object detection systems have a great potential impact for numerous applications, \emph{e.g.}, autonomous driving, robotics, and geometry quality inspection. Thanks to the rapid development of 3D acquisition technology, modern 3D object detection techniques attempt to tackle this task solely based on LiDAR point clouds derived from 3D sensors. Compared to 2D images, LiDAR point clouds retain rich geometry and shape information of objects, but the downside is that the irregular point clouds can not be processed via typical CNNs. This motivates a series of innovations to transform the irregular point clouds into the point-based \cite{qi2017pointnet, shi2019pointrcnn}, voxel-based \cite{yan2018second, zhou2018voxelnet}, or projection-based representation \cite{chen2017multi,ku2018joint}. After that, based on the representations, a surge of state-of-the-art 3D object detectors (\emph{e.g.}, PV-RCNN \cite{shi2020pv} and 3DSSD \cite{yang20203dssd}) have been constructed by adopting the generic single-stage \cite{liu2016ssd, lin2017focal} or two-stage 2D object detection \cite{ren2015faster,dai2016r} paradigm. However, considering that the input point clouds often suffer from sparsity due to long-distance or occlusion, directly leveraging the generic detection paradigms will inevitably result in unsatisfactory results, especially for objects with fewer/sparse points. Taking the 3D detection results in Figure \ref{fig:intro} as an example, the state-of-the-art 3D object detector (\emph{i.e.}, PV-RCNN) successes to localize the three cars with dense points, but fails to detect the other one car in the distance with extremely sparse points. On the contrary, our 3D Cascade RCNN manages to detect all the four cars.
	
	Such facts promote the development of recent 3D detection systems to alleviate the sparsity problem by combining more data/features from multiple sensors \cite{chen2017multi,liang2018deep,qi2018frustum} or involving additional components of point cloud refinement \cite{zhang2020pc}. The former commonly fuses LiDAR point clouds with RGB images to strengthen 3D object detection. However, the additional input RGB images are sensitive to illumination changes, which might make the final detection results less robust. The latter learns to enrich the points of target objects through point cloud completion. Such process of point cloud refinement leads to additional computation cost and is sometimes unsteady if the input points are deteriorated by occlusion. In contrast, we propose to mitigate these issues by re-weighting the task weight of each object proposal according to the sparsity level of points. This design automatically down-weights the object proposals with sparse points and thus mitigates their negative impacts over the whole training, without any auxiliary inputs or processing of point cloud refinement. Furthermore, it is well recognized that progressively improving proposal quality through a sequence of cascade detectors is crucial to high-quality object detections in 2D images. This naturally motivates us to exploit such completeness-aware re-weighting in the cascade paradigm, which iteratively refines the predicted 3D bounding boxes of proposals and further strengthens the confidences of completeness-aware re-weighting.

	By consolidating the idea of delving into the point cloud sparsity under the cascade paradigm, we present a new 3D Cascade RCNN for high-quality object detections tailored to point clouds. The whole architecture consists of a 3D backbone network, a sequence of cascade detection heads, and the Point Completeness Score module (PCS). Here the PCS module is specially devised to quantitatively measure the point completeness score for each proposal, which reflects the inherent sparsity level of the points within the 3D bounding box.
	Technically, the 3D backbone network is first leveraged to extract features for voxelized point clouds. After that, a sequence of detection heads is trained in a cascade fashion, in the sense that the output 3D proposals of a detection head are utilized to extract features for triggering the training of the next detection head. This cascade process gradually refines the output 3D proposals stage by stage. Meanwhile, for each output proposal, we capitalize on the PCS module to calculate the corresponding point completeness score, which is assigned as the task weight for training the detection head at each stage. Such designs of completeness-aware re-weighting strategy act as a key ingredient in 3D Cascade RCNN to prevent a cacophony of low-quality proposals with highly sparse points, yielding a more stable training process.

	In summary, we have made the following contributions:
	\begin{itemize}
		\item We quantitatively analyze the sparsity problem in 3D object detections and formalize a new definition of point completeness score to character the sparsity level of the points for each object. In particular, by capitalizing on the newly defined object sparsity, we provide a detailed quantitative analysis for sparsity problem, including the point completeness score distribution and the connections between sparsity and performance.
		\item To mitigate the negative effects over the whole training incurred by proposals with highly sparse points, we propose an effective (+1.1\% absolutely) and inference-friendly training mechanism, \emph{i.e.}, completeness-aware re-weighting strategy, that re-weights the task weight of each proposal according to its point completeness score.
		\item Aiming to further improve the detection of sparse objects, we devise a cascade architecture (\emph{i.e.}, 3D Cascade RCNN) that upgrades the generic cascade detection paradigm with a series of operations particularly for 3D object detection.
		\item The proposed 3D Cascade RCNN is extensively evaluated on KITTI dataset and Waymo Open Dataset, and obtains competitive performances. The source code is publicly available at \url{https://github.com/caiqi/Cascasde-3D}.
	\end{itemize}

	\section{Related Work}
	\subsection{3D Object Detection}
	The recent advances in large-scale point cloud datasets \cite{geiger2012we,sun2020scalability} and training deep neural networks \cite{he2016deep,qi2017pointnet,qi2017pointnet++,mopuri2018cnn,cotnet,cai2020joint} have inspired remarkable developments of 3D object detection \cite{bao2019monofenet,rahman2019notice,feng2020relation}. Generally, in analogy to 2D object detection, 3D object detection techniques can be briefly categorized into two directions: two-stage detectors based on 3D proposals \cite{chen2017multi,deng2020voxel,lahoud20172d,qi2019deep,qi2018frustum,shi2020pv,shi2020points,wang2019frustum,Yang_2019_ICCV} and one-stage detectors that directly applied on the primary points/voxels. \cite{engelcke2017vote3deep,He_2020_CVPR,lang2019pointpillars,yan2018second,yang2018pixor,yang20203dssd,zhou2018voxelnet}. In particular, \cite{lahoud20172d,qi2018frustum,wang2019frustum} first generate the 2D proposals from the RGB images, and PointNets \cite{qi2017pointnet} are further utilized to localize the objects in a frustum point cloud extruded from 2D proposals. PointRCNN \cite{shi2019pointrcnn} segments point clouds with PointNets to obtain foreground points, which are taken as the 3D proposals. Next, both semantic and local spatial features are leveraged to produce high-quality 3D boxes. STD \cite{Yang_2019_ICCV} further introduces a sparse-to-dense strategy for refining proposals. Instead of solely using point representations for detection, PV-RCNN \cite{shi2020pv} integrates the spatial context from voxel representation and the position information from raw points for each proposal. A RoI-pooling is further utilized to extract proposal-specific features for 3D object detection. \cite{deng2020voxel} upgrades \cite{shi2020pv} by directly exploiting 3D voxel tensors for detection in an efficient manner. Unlike two-stage detectors, one-stage approaches directly predict object categories and 3D box offsets without region proposal stage. For example, VoxelNet \cite{zhou2018voxelnet} divides point clouds into 3D voxels and capitalizes on PointNet to transform the points in each voxel into a unified feature representation. Then 3D CNN are leveraged to encode spatial context for 3D bounding box prediction. Second \cite{yan2018second} tackles the inefficiency of VoxelNet arising from sparsity of non-empty voxels by using the sparse 3D convolution networks. H$^2$3D R-CNN \cite{deng2021multi} proposes to fuse complementary information from perspective view and bird-eye view for 3D object detection. Some recent works \cite{He_2020_CVPR,lang2019pointpillars} further remould Single Shot Detector (SSD) \cite{liu2016ssd} to enable more efficient architectures for 3D object detection.

	\subsection{Sample Re-weighting}
	Sample re-weighting plays an essential role in training high quality object detection models. Classical object detectors (\emph{e.g.}, OHEM \cite{shrivastava2016training}) select hard samples during training and thus emphasize the optimization of the ``hard'' samples. Focal loss \cite{lin2017focal} discovers that the extreme foreground-background class imbalance is the central cause for low performances of one-stage detectors. A focal loss is thus proposed to down-weight the well-classified examples during training. Meanwhile, some works explore the opposite re-weighting directions to boost object detection. For example, Prime sampling \cite{cao2020prime} assigns larger weights to prime samples, which have higher IoUs with the ground-truth and are located more precisely around the object. IoU-balanced loss \cite{pang2019libra} identifies easy samples with the ranking of classification loss or the IoU with regard to ground truth, and assign higher weights for those ``easy'' samples. SWN \cite{cai2020learning} learns an adaptive sample re-weighting strategy based on the uncertainty of each prediction, and the high uncertainty (\emph{i.e.}, ``hard'') samples are down-weighted generally. Overall, these approaches determine the weighting of each sample solely based on the interaction (IoU, loss, etc.) between prediction and ground truth, while ignoring the intrinsic quality of the training sample. Some other recent works \cite{rezatofighi2019generalized,feng2021tood,zhang2021varifocalnet} on 2D detection fields use IoU score as a new regression target or include a branch for IoU-aware prediction. However, our PC Score is completely different from the IoU score in 2D detection framework. Specifically, the typical IoU score is measured as the IoU between the bounding boxes of ground truth and proposal, irrespective of the point cloud quality. Instead, our PC Score refers to the coverage ratio of the observed point clouds within the ground truth 3D bounding box, that reflects the inherent point cloud quality. As an alternative, we derive a re-weighting strategy tailored to 3D object detection, which is based on the inherent point cloud quality, and the subsequent experiments in Section \ref{sec:iou} demonstrate that such strategy exhibits better performances than IoU-based approaches.

	\subsection{Multi-Stage Object Detection}
	The past few years have witnessed the notable progress in 2D object detection, and the two-stage detectors \cite{chen2017multi,dai2016r,dai2017deformable,he2017mask,cai2020learning, cai2019exploring, deng2021minet} generally achieve better detection results than their one-stage counterparts \cite{redmon2016you,liu2016ssd,lin2017focal,deng2020single}. Motivated by the ``divide and conquer'' philosophy, the multi-stage detection architecture with cascade detectors \cite{cai2018cascade,chen2019hybrid,gidaris2015object,gidaris2016attend,yang2016craft} has attracted increasing attention and becomes a generic paradigm for high-quality object detection. Multi-Region CNN \cite{gidaris2015object} applies an R-CNN model multiple times that alternates between proposal scoring and localization refining. CRAFT \cite{yang2016craft} proposes to use cascade region networks for generating better proposals. A cascade of Fast RCNN is additionally utilized to reduce the false positives. Cascade RCNN \cite{cai2018cascade} designs a sequence of cascade detectors trained with increasing IoU thresholds. The outputs from each stage detector are fed into the next stage for iteratively refining proposals. R-FCN-3000 \cite{singh2018r} divides the classification step into a multi-stage framework, \emph{i.e.}, super-class classification and sub-class classification. The fine-grained category probability is obtained by multiplying the super-class probability with classification probabilities of sub-classes within each super-class. Such cascade paradigm can also be generalized to other tasks, \emph{e.g.}, object tracking \cite{fan2019siamese} and instance segmentation \cite{chen2019hybrid}.
	
	\subsection{Summary}
	Our 3D Cascade RCNN is also a type of cascade detection paradigm but tailored to 3D scenarios. Unlike Cascade RCNN that is developed for 2D object detection, ours goes beyond the iterative refinement of proposals, and additionally guides the cascade paradigm with the inherent sparsity level of each object to mitigate the sparsity problem. A completeness-aware re-weighting strategy is thus designed to screen out high-quality proposals at each stage for training, pursuing high-quality 3D object detection. Although cascade paradigm is exploited to progressively improve proposal quality in our approach for 3D object detection, the success of our work is more than that. In particular, our proposed 3D Cascade RCNN differs from existing techniques from two perspectives:
	\begin{itemize}
		\item Our work paves a new way to quantitatively analyze sparsity problems in 3D field by formulating a concrete definition of object sparsity (\emph{i.e.}, point completeness score) and provides a detailed quantitative analysis for this problem based on the newly defined object sparsity (see Sec.3 in the main paper), including the point completeness score distribution and the connections between sparsity and performance. To mitigate the negative effects over the whole training incurred by proposals with highly sparse points, we propose a simple, effective and inference-friendly training mechanism, \emph{i.e.}, completeness-aware re-weighting strategy, that re-weights the task weight of each proposal according to its point completeness score.
		\item The adaptation of cascade paradigm to 3D object detection is not trivial. We upgrade the typical cascade paradigm with a series of operations particularly for 3D object detection, \emph{e.g.}, fixed-IoU sampling strategy, voxel pooling+MLPs, and the reduced channel number in cascade detection head. Specifically, instead of the dynamic-IoU sampling strategy in 2D field that progressively enlarges IoU threshold, our 3D cascade paradigm utilizes a fixed-IoU sampling strategy at each stage, which is better aligned with the fixed-IoU evaluation metric in 3D field. Moreover, we adopt voxel pooling+MLPs to achieve proposal-level features, rather than the RoI pooling in 2D field. To further alleviate the additional computational cost caused by voxel pooling+MLPs, we slim the channels of RPN and RoI heads, leading to a better cost-performance tradeoff.
	\end{itemize}

	\section{Sparsity Problem in Point Clouds}
	
	It is well-known that the LiDAR point clouds in the real world are extremely sparse and the point density is always unbalanced tremendously. For example, by dividing the whole space into equal-sized voxels, VoxelNet \cite{zhou2018voxelnet} points out that more than 90\% voxels are empty. Owing to the sparsity of point clouds, most conventional 3D object detectors often suffer from sparsity problems and achieve low-quality detection results for objects with sparse points. Although the sparsity problem has been briefly mentioned in several existing works \cite{yan2018second,zhang2020pc,zhou2018voxelnet}, the problem of how the sparsity of objects affects the performances of 3D detectors is not yet fully understood in the literature. In this section, we take a close look at this problem and provide a detailed quantitative analysis, including the concrete definition of the sparsity for each object (named point completeness score), the score distribution of all objects, and the connections between sparsity and performance.
	
	\subsection{Point Completeness Score} To formalize this problem, we first present the concrete definition of point completeness score, that reflects the inherent sparsity level of each object. Intuitively, the objects with a large number of points are often easier to be detected, which can be treated as high-quality samples (see Figure \ref{fig:quality_examples} (b)). In this sense, one natural way to measure the point sparsity/quality for each object is directly using the number of observed points belonging to that object. Nevertheless, there might be a misalignment between point number and point cloud quality sometimes. As Figure \ref{fig:quality_examples} (c) illustrates, despite the point number for this object is large, most of the points are locally clustered and thus fail to cover the complete outline of the target object, thereby resulting in a low-quality sample. Motivated by this, for each object, we measure the point completeness score as the coverage ratio of its observed points clouds within the ground truth 3D bounding box, which can be interpreted as the object point quality for detection.
	
	Formally, suppose we are given the ground truth 3D bounding box $B=(x,y,z,l,w,h,r)$ of an object that contains $N$ observed points $\mathcal{P} = \{p_1, p_2,..., p_N\}$. $(x,y,z)$ and $(l,w,h)$ represents the center coordinates and size in X-axis, Y-axis, Z-axis, respectively. $r$ denotes the rotation angle along Z-axis. By denoting $A = (\hat{x}, \hat{y}, \hat{z}, \hat{l}, \hat{w}, \hat{h}, r)$ as the smallest enclosing box aligned with $B$ that encloses all points in $\mathcal{P}$, the point completeness score (PC Score) is calculated as the IoU between $A$ and $B$:

	\begin{figure}[!tb]
		\centering
		\begin{tabular}{cc}
			\includegraphics[height=15mm]{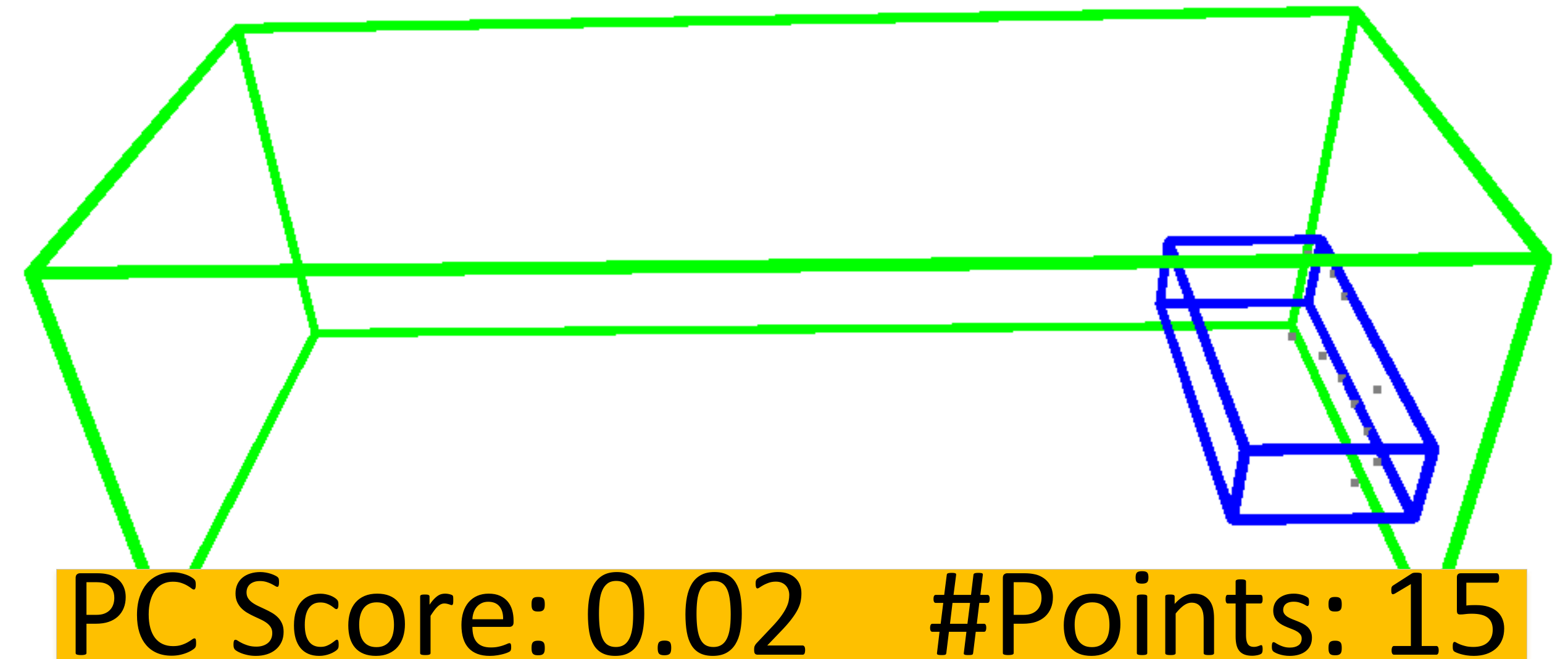} & \includegraphics[height=15mm]{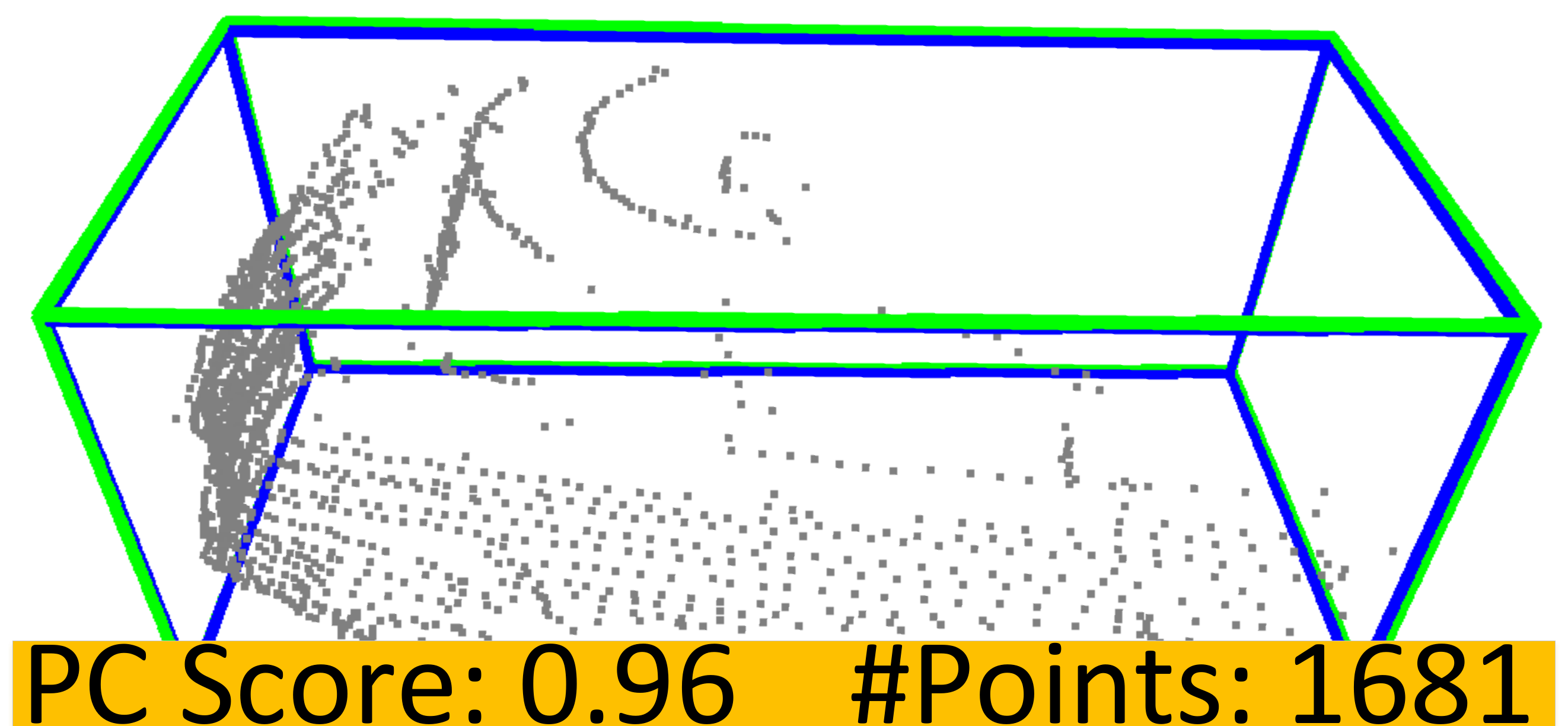} \\
			(a)                                            & (b)                                            \\
			\includegraphics[height=15mm]{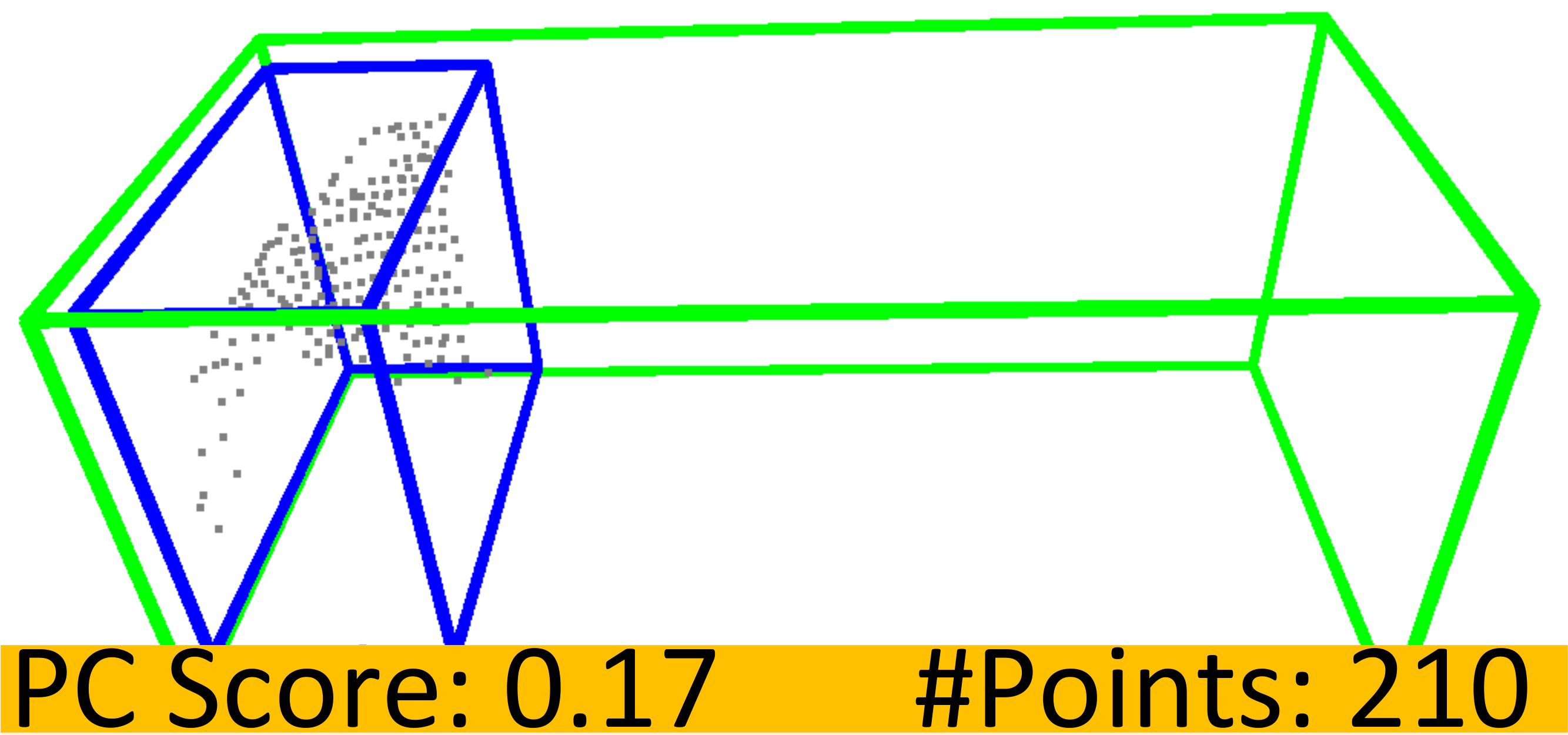} & \includegraphics[height=15mm]{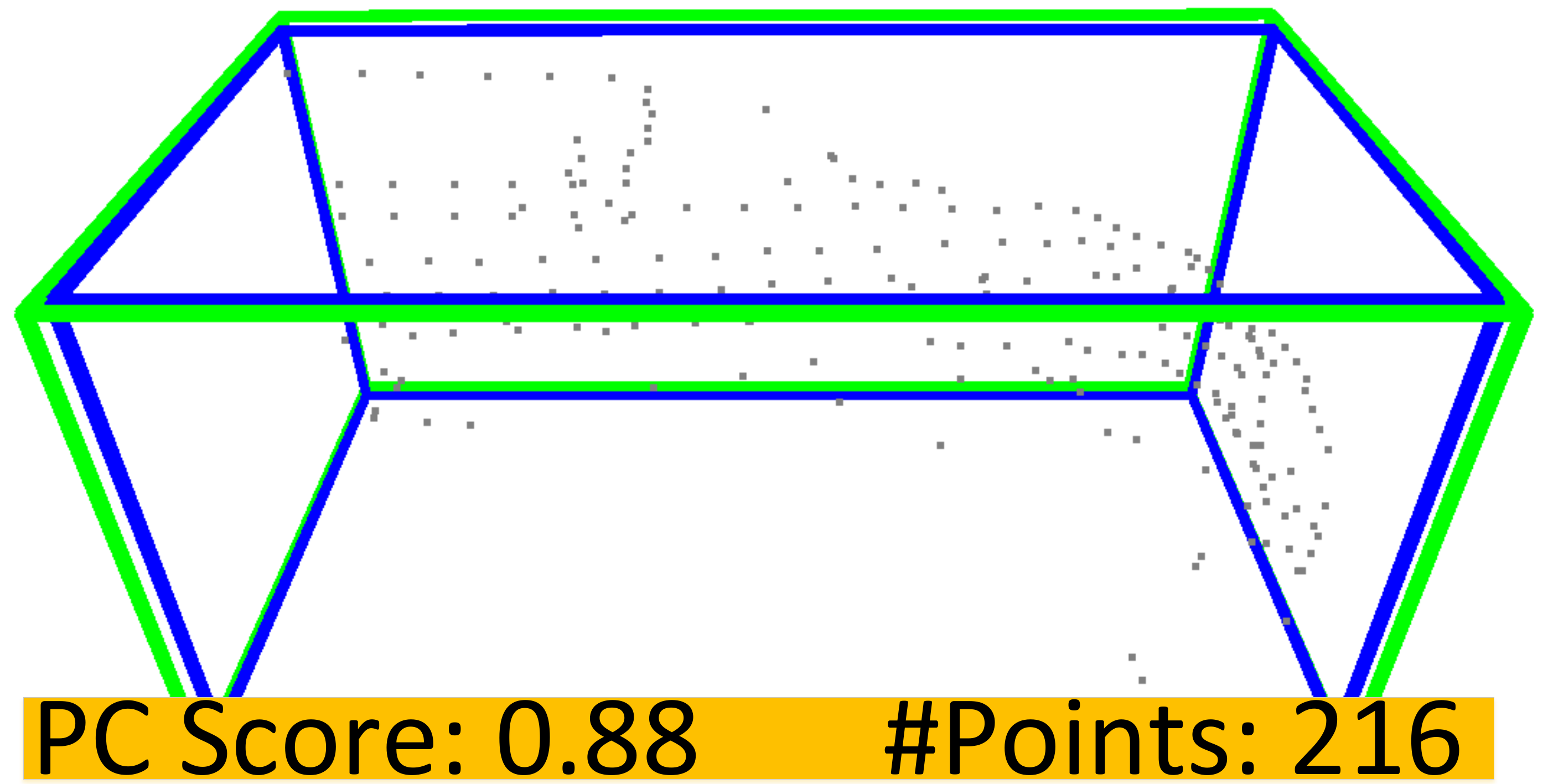} \\
			(c)                                            & (d)                                            \\
		\end{tabular}
		\caption{\small Example point clouds of objects in KITTI dataset. \textcolor{green}{GREEN} denotes the ground truth bounding box and \textcolor{blue}{BLUE} is the smallest enclosing box of points. (a)\&(b): The objects having few/large number of points and low/high point completeness score. (c)\&(d): The samples exhibiting inconsistency between point number and point completeness score.}
		\label{fig:quality_examples}
	\end{figure}

	\begin{equation}\label{eq:quality_defination}
		\small
		\begin{aligned}
			Q & = \frac{A \cap B }{A \cup B }.
		\end{aligned}
	\end{equation}
	Note that A is the smallest enclosing box of points within the ground truth box B. Thus, the union of A and B is identical to B and Eq. (\ref{eq:quality_defination}) can be written as:
	\begin{equation}\label{eq:quality_defination_simplified}
		\small
		\begin{aligned}
			Q & = \frac{A \cap B }{ B }.
		\end{aligned}
	\end{equation}

	\subsection{PC Score Distribution} 
\begin{figure}[!tb]
		\centering {\includegraphics[width=0.5\textwidth]{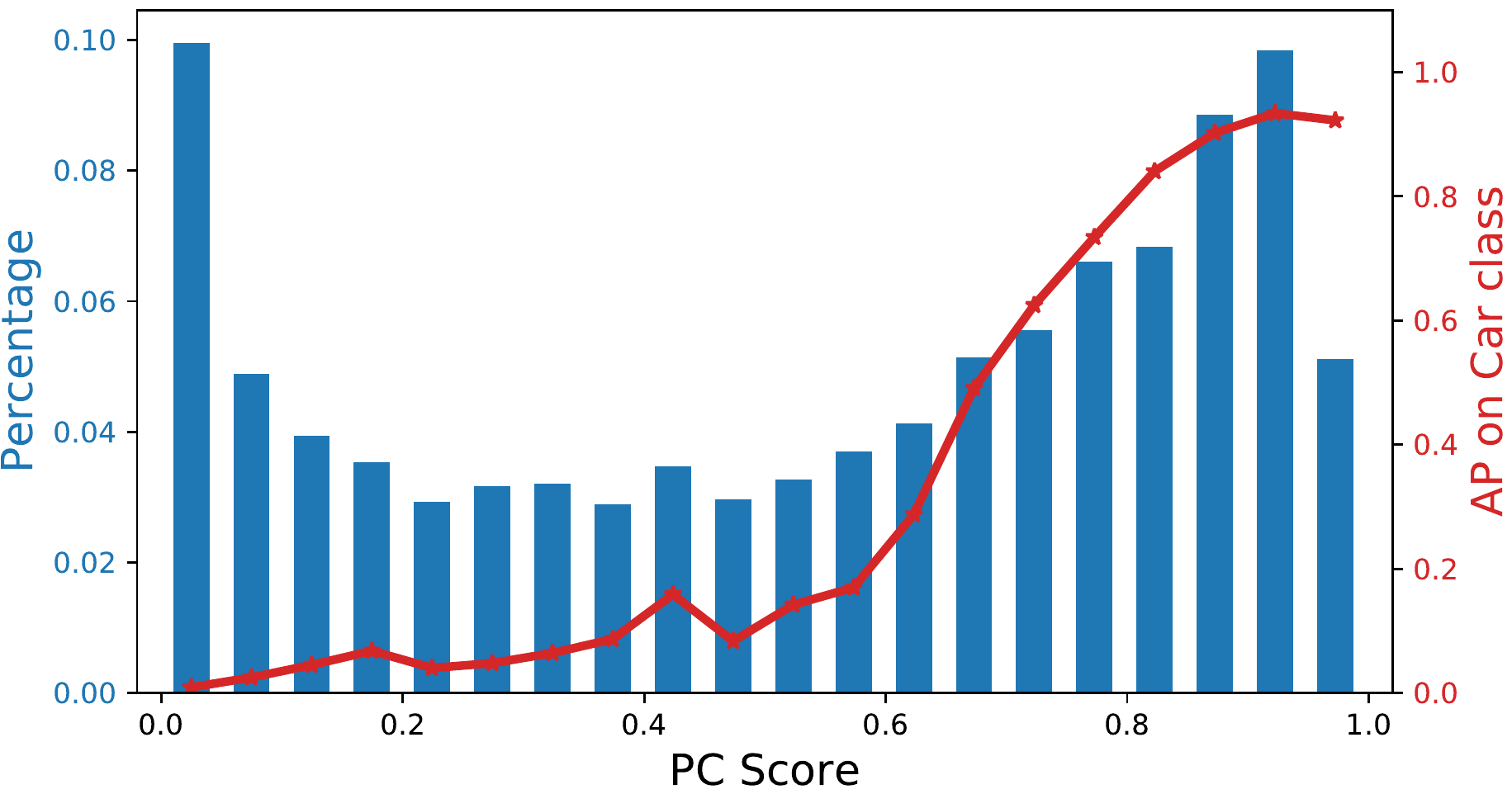}}
		\caption{\small Point completeness score distribution (in \textcolor{blue}{BLUE}) and performance curve for objects with different point completeness scores (in \textcolor{red}{RED}) on KITTI \textit{val-set}. The AP is computed by ignoring ground truth whose PC score out of specific bin range.
		}
		\label{fig:gt_quality}
	\end{figure}

After clearly defining the sparsity of each object as the point completeness score, we quantitatively analyze the distribution of point completeness scores of all samples derived from KITTI \textit{val-set} (Figure \ref{fig:gt_quality}). Specifically, around 10\% of objects achieve very low point completeness scores (below 0.05) .
	Moreover, we find that the objects with sparse points (the PC Score $Q \textless 0.5$) account for a large proportion (more than 50\%) of samples. These observations quantitatively demonstrate that the sparsity of point clouds is a common issue in existing 3D detection datasets.

	\begin{figure*}[!tb]
		\centering {\includegraphics[width=1.0\textwidth]{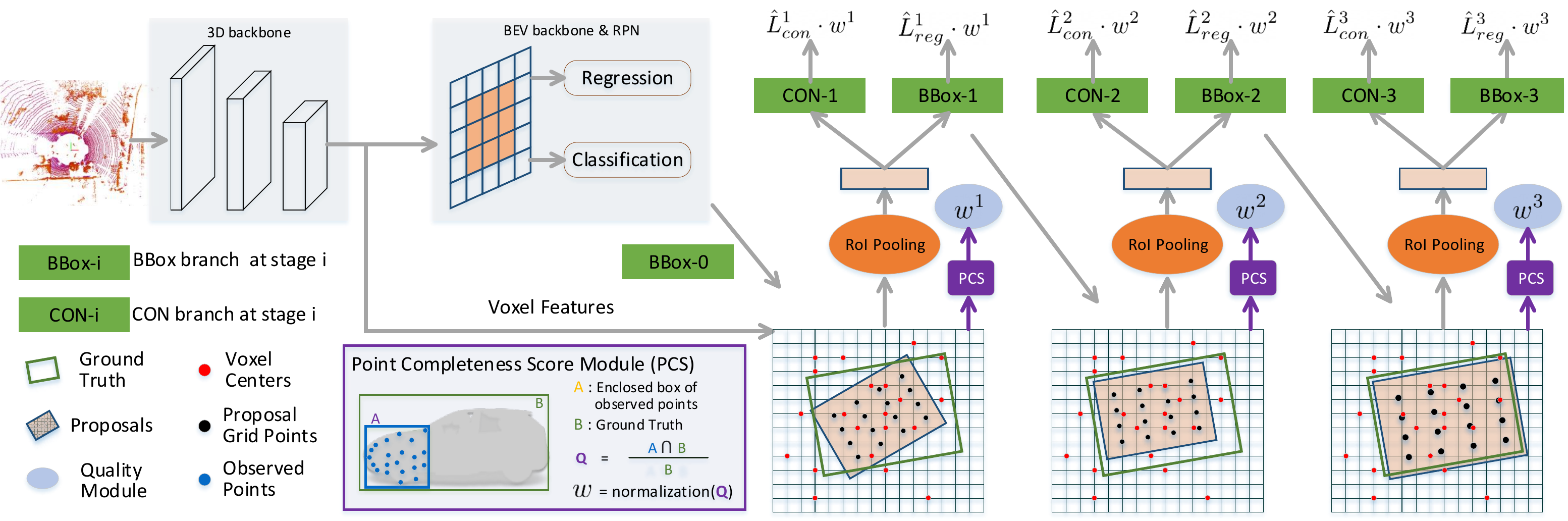}}
		\caption{\small Overview of our 3D Cascade RCNN for 3D object detection. 3D Backbone Network first encodes the voxelized point clouds into 3D feature volumes. Next, the 3D feature volumes are stacked along Z-axis to compose the Bird-Eye-View (BEV) representation, which is fed into RPN for initial proposal generation. Finally, the cascade detection heads iteratively refine proposals and predict confidences in a cascade paradigm. Meanwhile, for each stage, we utilize the Point Completeness Score module (PCS) to measure the point completeness score of each proposal, which triggers the completeness-aware re-weighting that re-weight the task weight of each proposal during training.}
		\label{fig:framework}
	\end{figure*}
	
	\subsection{Performance Curve w.r.t Point Completeness Score} We further investigate the connections between the detection performance and the point completeness score by measuring the average performance (Moderate AP11 of Car) of object samples with similar point completeness score in KITTI \textit{val-set}. Note that here we adopt PV-RCNN \cite{shi2020pv} as the base detector. As shown in the performance curve in Figure \ref{fig:gt_quality}, when the point quality of object becomes worse with a lower point completeness score, the detection performance is gradually decreased. We speculate that this may be the result of the cacophony of low-quality objects with highly sparse points, which poses a severe challenge for existing generic detectors. Moreover, the low-quality objects could also deteriorate the training process by dominating the overall objective with larger losses.

	\section{3D Cascade RCNN} \label{sec:3dcascadercnn}
	Now we proceed to present our central proposal, \emph{i.e.}, 3D Cascade RCNN, that remolds the generic cascade 2D detection paradigm for 3D object detection. Under such a cascade paradigm, we further delve into the sparsity problem by introducing a new completeness-aware re-weighting strategy at each stage, which targets high-quality detection without increasing any FLOP budget. The detailed architecture of our 3D Cascade RCNN is illustrated in Figure \ref{fig:framework}.
	
	\subsection{Preliminaries}
	We first provide a brief review of the typical two-stage 3D detector (Voxel R-CNN \cite{deng2020voxel}), which is composed of 3D Backbone Network, Region Proposal Network (RPN), and one Detection Head.
	
	\textbf{3D Backbone Network.} Given the input point cloud with the scale of ($W$, $H$, $D$) along the X-axis, Y-axis, and Z-axis, respectively, it is firstly divided into equally spaced 3D voxels. By defining the voxel size as (${v}_W$, ${v}_H$, ${v}_D$), the dimension of each voxel grid is thus represented as ($W'=W/{v}_W$, $H'=H/{v}_H$, $D'=D/{v}_D$). After that, the 3D backbone network \cite{yan2018second} employs several stacked 3D sparse convolutions to convert the voxels into progressively downsampled ($1\times$, $2\times$, $4\times$, $8\times$) 3D feature volumes.
	
	\textbf{Region Proposal Network.} The 3D feature volumes are further compressed along Z-axis into the Bird-Eye-View (BEV) representation. Next, based on the BEV representation, a 2D backbone consisting of standard convolution with non-linearity is applied to further encode the features. Two sibling $1 \times 1$ convolutions are appended on top of the 2D backbone to produce 3D region proposals with the category and bounding box offsets predictions.

	\textbf{Detection Head.} The detection head is utilized to extract proposal-specific features from 3D feature volumes via RoI pooling according to the generated proposals from RPN. Specifically, we leverage RoI-pooling to aggregate features for each grid point, which are further encoded with several MLPs into proposal-level features. Finally, conditioned on the proposal-specific features, confidence (CON) and regression (BBox) branches are applied to produce the confidence score and bounding box offsets, respectively.
	
	\begin{figure}[!tb]
		\centering {\includegraphics[width=0.4\textwidth]{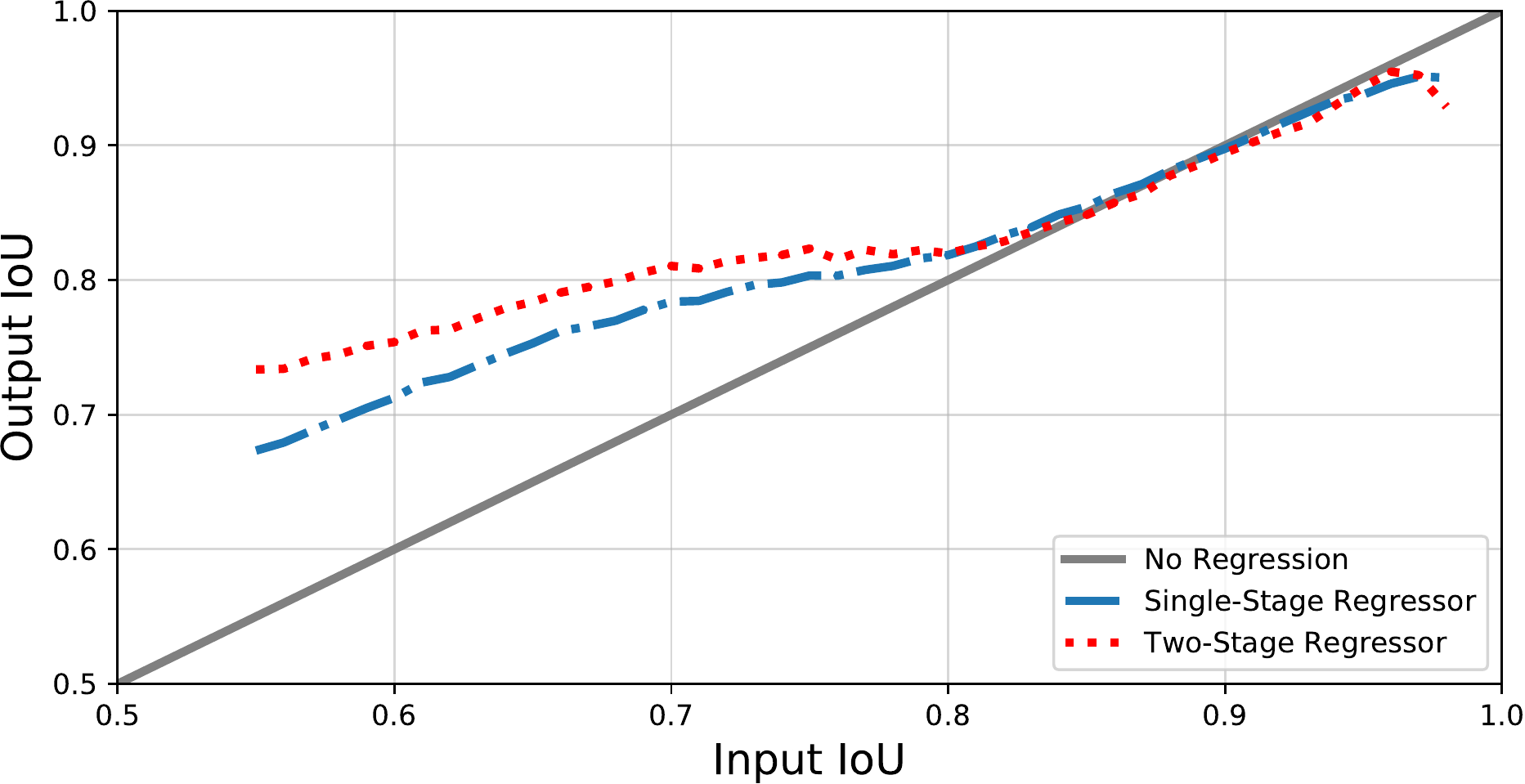}}
		\caption{\small Localization performance of 3D object detectors with single-stage and two-stage bounding box regressor. Two-stage regressor exhibits consistently better performances than single-stage regressor when the input IoU is less than 0.8.}
		\label{fig:iou_gains}
	\end{figure}
	
	\subsection{Cascade Detection Heads}\label{sec:cascade_roi_head}
	Most of existing two-stage 3D object detectors (\emph{e.g.}, STD \cite{Yang_2019_ICCV} and PV-RCNN \cite{shi2020pv}) are constructed with a single detection head, that regresses 3D proposals only once. Nevertheless, several 2D detection techniques \cite{cai2018cascade,gidaris2015object,gidaris2016attend} have argued that a single detection head reflects limited localization capacity and is hardly applicable to all proposals at different quality levels. Inspired by \cite{cai2018cascade} that performs iterative proposal refinement via cascade detectors in 2D images, we derive a particular form of cascade detection heads to mitigate such limitation in 3D scenarios. Technically, we decompose the typical single detection head into $T$ consecutive detection heads in multi-stage manner. Here the detection head in each stage consists of a regression branch $g^b_t$ and confidence branch $g^c_t$ where $t \in \{1, 2,...,T\}$. We take the region proposals from RPN as the stage-0 proposals $R_0 = \{r^0_0 , r^0_1, ..., r^0_M\}$, and the refined proposals from $t$-th stage are denoted as $R_t = \{r^t_0 , r^t_1, ..., r^t_M\}$. Here $M$ is the number of total proposals. In $t$-th stage, RoI pooling is utilized to extract RoI feature $f^t_m$ for each proposal $r^t_m$:
	\begin{equation}\label{eq:voxel_pooling}
		\small
		\begin{aligned}
			f^t_m & = \text{RoI-Pooling}( F_{3D}, r^{t-1}_m), m \in \{1, ..., M\},
		\end{aligned}
	\end{equation}
	where $F_{3D}$ is the 3D feature volumes derived from 3D backbone network.
	Next, given the RoI feature $f^t_m$, the confidence branch $g^c_t$ and regression branch $g^b_t$ produces the confidence scores and refined proposals, respectively:
	\begin{equation}\label{eq:roi_head_prediction}
		\small
		\begin{aligned}
			c^t_m  = g^{c}_t( f^t_m ), r^t_m = g^{b}_t( f^t_m ).
		\end{aligned}
	\end{equation}
	Eq. (\ref{eq:voxel_pooling}) and Eq. (\ref{eq:roi_head_prediction}) are iteratively performed until the final stage. At inference, we perform mean pooling over the predicted confidence scores of all stages as the final confidence prediction, and the refined proposal of last stage is directly taken as the output 3D bounding box:
	\begin{equation}\label{eq:inference_output}
		\small
		\begin{aligned}
			c_m =\frac{1}{T} \sum_{t=1}^T  c^t_m,  r_m = r^T_m.
		\end{aligned}
	\end{equation}
	Here we further conduct the localization performance comparison between single-stage and two-stage 3D bounding box regressors on KITTI \textit{val-set}. Note that the two-stage regressor refers to the regressor in our cascade detection heads (\emph{i.e.}, $T=3$). As depicted in Figure \ref{fig:iou_gains}, the two-stage regressor in a cascade manner consistently outperforms the single-stage regressor when the input IoU is less than 0.8. The results basically validate our hypothesis that compared to single-stage detection techniques, the cascade detection paradigm can produce higher quality 3D bounding boxes through iterative proposal refinement.
	
	\subsection{Completeness-Aware Re-weighting} \label{sec:pcs}
	
	The most typical way to train the detection head at each stage is to directly fuse the confidence and regression losses from two branches as in the generic detection paradigm, while ignoring the point sparsity/quality unbalance across all objects. This inevitably incurs larger losses for a large number of low-quality samples, which overwhelms the relatively small losses of high-quality samples and results in an unsteady training process. Instead, we design a novel completeness-aware re-weighting strategy to re-weight the task weight of each proposal according to its point completeness score. The rationale behind is to reshape the overall training objective by down-weighting the low-quality samples, and thus focus training on high-quality samples. The low-quality samples will incur large losses, which overwhelm the relatively small losses of high-quality samples. Such unbalanced loss distribution inevitably results in an unsteady training process. Instead, our re-weighting strategy down-weights the low-quality samples, pursuing a more balanced loss distribution.

	Formally, for each region proposal $r^t_m$ at $t$-th stage, we calculate its task weight $w^t_m$ depending on the point completeness score of the corresponding matched ground truth object. The task weights of proposals which cannot be matched to any ground truth object remain unchanged. More specifically, we first capitalize on Point Completeness Score module (PCS) to achieve the point completeness score $Q^t_m$ of the matched ground truth of $r^t_m$ via Eq. (\ref{eq:quality_defination}). Note that $Q^t_m$ is only valid for positive proposals $\mathcal{R}_t^P$ and set to 0 for negative proposals. Next, a linear normalization function is employed to obtain the task weight:
	\begin{equation}
		\small
		\begin{aligned}
			\hat{w}^t_m =  |\mathcal{R}_t^P| \cdot \frac{Q^t_m}{ \sum_{j=1}^M Q^t_j}
		\end{aligned}\label{eq:linear_task_weight}
	\end{equation}
	$|R_t^p|$ denotes the number of positive proposals. We regard it as a normalization factor to ensure that the sum of all task weights for positive samples stays the same before and after re-weighting. Specifically, when we do not apply the re-weighting strategy, the sum of task weights for all positive samples is $|R_t^p|$ (\emph{i.e.}, the weight for each positive sample is set as 1). After applying the re-weighting strategy, if we remove this normalization factor, the sum of task weights for all positive samples is 1, which is inconsistent with the scale before re-weighting. The linear normalization function can also be replaced with Softmax function, but we did not observe improved performance in the experiments. Please note that the re-weighting strategy is only applied to proposals which can be matched with ground truth boxes (\emph{i.e.}, the positive proposals $\mathcal{R}_t^P$). Accordingly, the task weights for all proposals at each stage are calculated as:
	\begin{equation}
		w^t_m = \begin{cases}
			\hat{w}^t_m & r^t_m \in \mathcal{R}_t^P, \\
			1           & \text{otherwise}.
		\end{cases} \label{eq:final_task_weights}
	\end{equation}
	
	\begin{figure}[!tb]
		\centering {\includegraphics[width=0.46\textwidth]{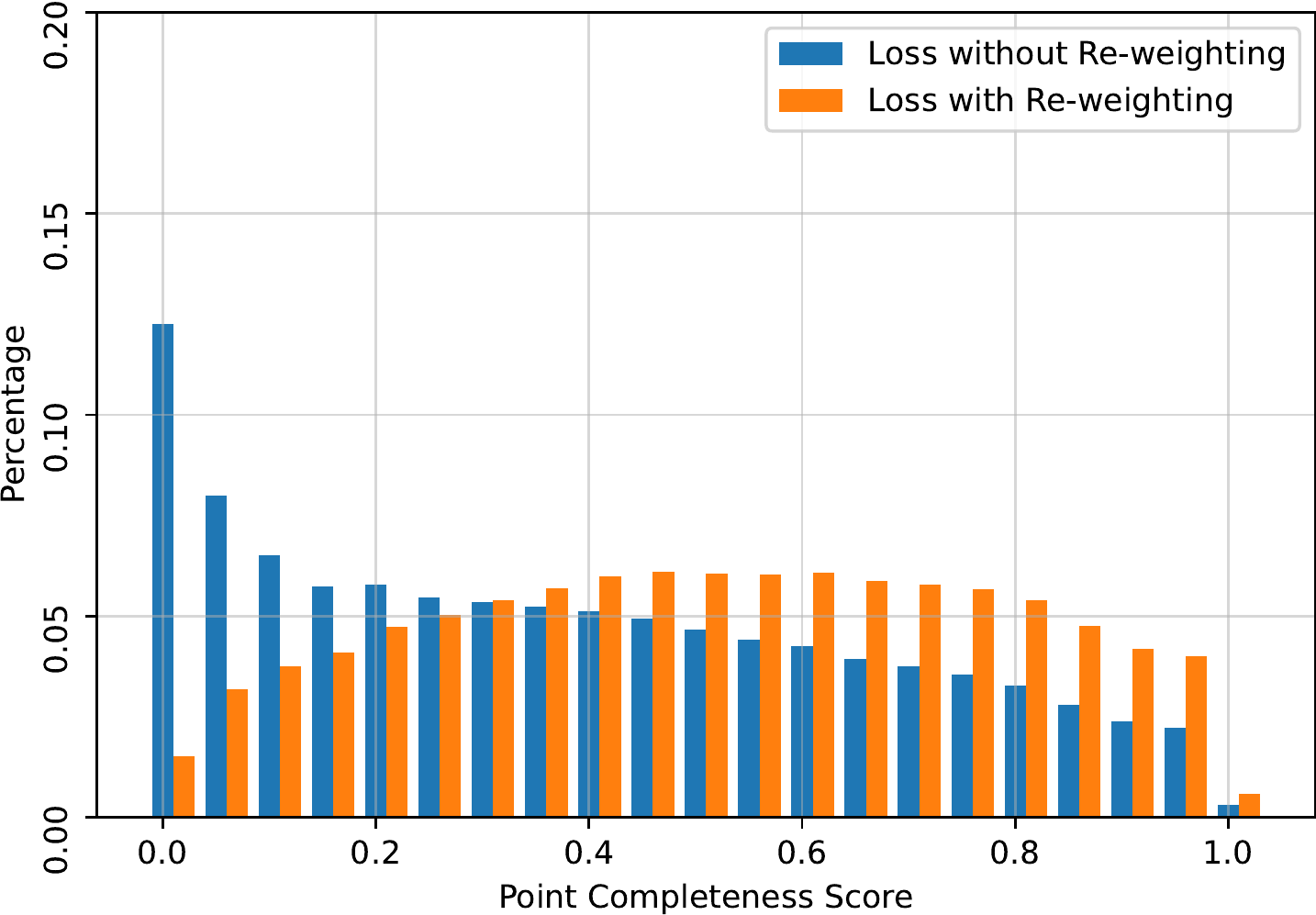}}
		\caption{\small Loss distributions of different Point Completeness Scores with and without re-weighting.}
		\label{fig:loss_distribution}
	\end{figure}
	
	We visualize the loss distribution changes of the proposed re-weighting strategy in Figure \ref{fig:loss_distribution}. Note that although the weights of low-quality samples have been reduced, the losses of them (PC score$\in[0.2, 0.5]$) still contribute a similar portion with the ones of high-quality samples (PC score$\in[0.5, 0.9]$). This confirms our hypothesis that the completeness-aware re-weighting enables a more steady training process that equally optimizes most samples across different qualities. It is worthy to note that similar observations are also reported in 2D field like Prime Sample \cite{cao2020prime} and Sample Weighting Network \cite{cai2020learning}. For example, Prime Sample assigns larger weights to prime samples, which have higher IoUs with the ground-truth and are located more precisely around the object. Sample Weighting Network proposes to draw more attentions to “easy” samples (\emph{i.e.}, the ones with high IoU values).
	
	\subsection{Training Objective}\label{sec:objective}
	The overall training objective of 3D Cascade RCNN is composed of losses in RPN and cascade detection heads. For the objective of RPN, we follow \cite{shi2020pv} to measure it as the fusion of classification loss $L_{rpn}^{cls}$ (as in Focal loss \cite{lin2017focal}) and Smooth-L1 loss $L_{rpn}^{reg}$. The objective of each detection head consists of two components: the IoU-guided confidence loss $\hat{L}_{con\text{-}m}^t$ \cite{shi2020pv} implemented as Binary Cross Entropy and the regression loss $\hat{L}_{reg\text{-}m}^t$ (\emph{i.e.}, Smooth-L1 loss). Given the task weights defined as in Eq. (\ref{eq:final_task_weights}), the objective of each detection head at $t$-th stage is thus measured as:
	\begin{equation} \label{eq:rcnn_loss_single_stage}
		\small
		\begin{aligned}
			L_{rcnn\text{-}m}^t & = w^t_m \cdot \hat{L}_{con\text{-}m}^t +  w^t_m \cdot \hat{L}_{reg\text{-}m}^t.
		\end{aligned}
	\end{equation}
	The objective of cascade detection heads is finally calculated by aggregating the losses of all stages:
	\begin{equation} \label{eq:rcnn_loss}
		\small
		\begin{aligned}
			L_{rcnn\text{-}m} & = \sum_{t=1}^T L_{rcnn\text{-}m}^t.
		\end{aligned}
	\end{equation}
	
	\section{Experiments}
	We evaluate our 3D Cascade RCNN on the widely adopted KITTI \cite{geiger2012we} dataset and the recently released large-scale Waymo Open Dataset \cite{sun2020scalability} for 3D object detection.
	\subsection{Dataset and Implementation Details}
	
	\textbf{KITTI Dataset} is a popular dataset for benchmarking 3D object detection techniques in outdoor scenes. We follow the common settings in PV-RCNN \cite{shi2020pv} by splitting the 7,481 training samples (\textit{trainval-set}) into \textit{train-set} (3,712 samples) and \textit{val-set} (3,769 samples). The official test split \textit{test-set} contains 7,518 samples without disclosed labels. Therefore, we not only report the performances on \textit{val-set}, but also submit the results to online test server for \textit{test-set} evaluation. For experiments on the \textit{val-set}, we train models on the \textit{train-set} and for \textit{test-set} evaluation, the models are trained over \textit{trainval-set} as in \cite{shi2020pv}. We adopt average precision (AP) with IoU threshold of 0.7 for cars as evaluation metric. The official calculation of AP is changed from 11 recall positions to 40 recall positions on the day of 08.10.2019\footnote{\url{http://www.cvlibs.net/datasets/kitti/eval_object.php?obj_benchmark=3d}}. Thus we report performances with the AP setting of 40 recall positions (AP40) for both \textit{val-set} and \textit{test-set}. The model is additionally evaluated with the AP of 11 recalls (AP11) on \textit{val-set} to compare with previous approaches.
	
	\textbf{Waymo Open Dataset} is a recently released large-scale dataset, that is 15$\times$ larger than KITTI. It consists of 798 sequences for training and 202 sequences for testing. Each sequence contains multiple point cloud samples, leading to around 158k training samples and 40k validation samples. Compared with KITTI that only annotates objects in the camera field of view, Waymo Open Dataset provides annotations in the full 360\degree field. We adopt the official evaluation code\footnote{\url{https://github.com/waymo-research/waymo-open-dataset}} to report the performances. Specifically, mean average precision (mAP) in 3D and BEV views are used and the IoU threshold is set as 0.7 for vehicle detection. The test data is split into different difficulty levels based on the object's distance from the sensor (0-30m, 30-50m, 50m-inf) and the number of observed points within objects (LEVEL\_1 $\ge$ 5 points and LEVEL\_2 $\ge$ 1 point).
	
	\textbf{Point Cloud Voxelization.}
	On KITTI, the range of point clouds is clipped inside [0, 70.4]m along X-axis, [-40,40]m along Y-axis, and [-3, 1]m along Z-axis. The voxel size is (0.05, 0.05, 0.1)m in X/Y/Z-axis, yielding a voxel grid with size (1,408, 1,600, 40). For Waymo Open Dataset, the detection range is [-75.2, 75.2]m along X and Y axes, and [-2, 4]m for Z-axis. The voxel size is set as (0.1, 0.1, 0.15)m.
	
	\textbf{Network Structure.}
	For KITTI, as in \cite{shi2020pv}, the 3D backbone network is constructed with four stages having filter sizes of [16, 32, 64, 64], and the stride is 2 from the second stage. The resolution of final output 3D features is 8$\times$ down-sampled. The backbone for RPN contains two blocks with filter size [64, 96]. The RoI-pooling extracts features from 3-th and 4-th stages in 3D backbone network. For Waymo Open Dataset, we set the filter size for the RPN network as [128, 256].  The cascade detection heads consist of three stages.
	
	\begin{table}[!t]
		\small
		\renewcommand\arraystretch{1}
		\begin{center}
			\setlength\tabcolsep{4pt}
			\begin{tabular}{c|c|ccc}
				\hline
				\multirow{2}{*}{Method}                  & \multirow{2}{*}{Reference} &
				\multicolumn{3}{c}{$\text{AP}_\text{3D}$ (\%) ~~}                                                                                           \\
				&                            & Easy           & Moderate                          & Hard           \\
				\hline
				VoxelNet \cite{zhou2018voxelnet}         & CVPR 2018                  & 81.97          & \cellcolor{blue!10}65.46          & 62.85          \\
				SECOND \cite{yan2018second}              & Sensors 2018               & 88.61          & \cellcolor{blue!10}78.62          & 77.22          \\
				PointPillars \cite{lang2019pointpillars} & CVPR 2019                  & 86.62          & \cellcolor{blue!10}76.06          & 68.91          \\
				PointRCNN \cite{shi2019pointrcnn}        & CVPR 2019                  & 88.88          & \cellcolor{blue!10}78.63          & 77.38          \\
				3DSSD \cite{Yang_2020_CVPR}              & CVPR 2020                  & 89.71          & \cellcolor{blue!10}79.45          & 78.67          \\
				Part-$A^2$ \cite{shi2020points}          & TPAMI 2020                 & 89.47          & \cellcolor{blue!10}79.47          & 78.54          \\
				TANet \cite{liu2020tanet}                & AAAI 2020                  & 87.52          & \cellcolor{blue!10}76.64          & 73.86          \\
				PV-RCNN \cite{Shi_2020_CVPR}             & CVPR 2020                  & 89.35          & \cellcolor{blue!10}83.69          & 78.70          \\
				CIA-SSD \cite{zheng2020cia}              & AAAI 2021                  & 89.59          & \cellcolor{blue!10}80.28          & 72.87          \\
				Voxel R-CNN \cite{deng2020voxel}         & AAAI 2021                  & 89.41          & \cellcolor{blue!10}84.52          & 78.93          \\	\hline \hline
				3D Cascade RCNN                          & -                          & \textbf{90.05} & \cellcolor{blue!10}\textbf{86.02} & \textbf{79.27} \\	\hline
			\end{tabular}
		\end{center}
		\caption{\small Performance comparison on KITTI \textit{val-set} with AP of 11 recall positions (AP11) for car class.}
		\label{tab:val_r11}
	\end{table}
	
	\begin{table}[!t]
		\small
		\begin{center}
			\setlength\tabcolsep{2pt}
			\begin{tabular}{c|ccc|ccc}
				\hline
				\multirow{2}{*}{Method}         & \multicolumn{3}{c|}{$\text{AP}_\text{3D}$ (\%) ~~} & \multicolumn{3}{c}{$\text{AP}_\text{BEV}$ (\%) ~~}                                                                                        \\
				& Easy                                               & Moderate                                           & Hard           & Easy           & Moderate                          & Hard           \\
				\hline
				PV-RCNN \cite{shi2020pv}        & 92.57                                              & \cellcolor{blue!10}84.83                           & 82.69          & 95.76          & \cellcolor{blue!10}91.11          & 88.93          \\
				Voxel R-CNN \cite{deng2020voxel} & 92.38                                              & \cellcolor{blue!10}85.29                           & 82.86          & 95.52          & \cellcolor{blue!10}91.25          & 88.09          \\
				3D Cascade RCNN                 & \textbf{93.20}                                     & \cellcolor{blue!10}\textbf{86.19}                  & \textbf{83.48} & \textbf{96.31} & \cellcolor{blue!10}\textbf{92.41} & \textbf{89.95} \\ \hline
			\end{tabular}
		\end{center}
		\caption{\small Performance comparison on KITTI \textit{val-set} with AP and BEV AP of 40 recall positions (AP40) for car class.}
		\label{tab:kitti_val_ap40}
	\end{table}
	
	\textbf{Training and Inference.}
	Our 3D Cascade RCNN is trained with Adam \cite{kingma2014adam} optimizer in an end-to-end manner. We set the initial learning rate as 0.01, which is decayed with a cosine schedule. For KITTI dataset, the batch size is 16 and we train the whole model for 80 epochs. For Waymo Open Dataset, we set the batch size as 8 and train the model for 30 epochs. The foreground IoU threshold for regression is set as 0.55 for all stages on both datasets. For each stage, we sample 128 RoIs to train the detection head. Following \cite{shi2020pv}, we utilize the widely adopted data augmentations, including random flipping along X-axis, global scaling with random scale factor in [0.95, 1.05], global rotating point clouds along Z-axis by angles in [-$\frac{\pi}{4}$, $\frac{\pi}{4}$], and ground-truth sampling. At inference, the region proposals are non-maximum suppressed (NMS) with IoU threshold 0.7, and the top-100 proposals are fed into the detection head. Finally,  NMS with IoU threshold 0.1 is applied to further reduce duplicate detection results.

	\subsection{Performance Comparisons}

	\begin{table}[!t]
		\small
		\renewcommand\arraystretch{1.05}
		\begin{center}
			\setlength\tabcolsep{4pt}
			\begin{tabular}{c|c|ccc}
				\hline
				\multirow{2}{*}{Method}                  &
				\multirow{2}{*}{Reference}               &
				\multicolumn{3}{c}{$\text{AP}_\text{3D}$ (\%) ~~}                                                                             \\
				&              & Easy           & Mod.                              & Hard           \\
				\hline
				MV3D \cite{chen2017multi}                & CVPR 2017    & 74.97          & \cellcolor{blue!10}63.63          & 54.00          \\
				F-PointNet \cite{qi2018frustum}          & CVPR 2018    & 82.19          & \cellcolor{blue!10}69.79          & 60.59          \\
				AVOD-FPN \cite{ku2018joint}              & IROS 2018    & 83.07          & \cellcolor{blue!10}71.76          & 65.73          \\
				PointSIFT+SENet \cite{zhao20193d}        & AAAI 2019    & 85.99          & \cellcolor{blue!10}72.72          & 64.58          \\
				UberATG-MMF \cite{liang2019multi}        & CVPR 2019    & 88.40          & \cellcolor{blue!10}77.43          & 70.22          \\
				VoxelNet \cite{zhou2018voxelnet}         & CVPR 2018    & 77.47          & \cellcolor{blue!10}65.11          & 57.73          \\
				SECOND \cite{yan2018second}              & Sensors 2018 & 83.34          & \cellcolor{blue!10}72.55          & 65.82          \\
				PointPillars \cite{lang2019pointpillars} & CVPR 2019    & 82.58          & \cellcolor{blue!10}74.31          & 68.99          \\
				PointRCNN \cite{shi2019pointrcnn}        & CVPR 2019    & 86.96          & \cellcolor{blue!10}75.64          & 70.70          \\
				TANet \cite{liu2020tanet}                & AAAI 2020    & 85.94          & \cellcolor{blue!10}75.76          & 68.32          \\
				HVNet \cite{Ye_2020_CVPR}                & CVPR 2020    & 87.21          & \cellcolor{blue!10}77.58          & 71.79          \\
				Part-$A^2$ \cite{shi2020points}          & TPAMI 2020   & 87.81          & \cellcolor{blue!10}78.49          & 73.51          \\
				3DSSD \cite{Yang_2020_CVPR}              & CVPR 2020    & 88.36          & \cellcolor{blue!10}79.57          & 74.55          \\
				SA-SSD \cite{He_2020_CVPR}               & CVPR 2020    & 88.75          & \cellcolor{blue!10}79.79          & 74.16          \\
				PV-RCNN \cite{Shi_2020_CVPR}             & CVPR 2020    & 90.25          & \cellcolor{blue!10}81.43          & 76.82          \\
				Voxel R-CNN \cite{deng2020voxel}         & AAAI 2021    & \textbf{90.90} & \cellcolor{blue!10}81.62          & 77.06          \\
				\hline\hline
				3D Cascade RCNN                          & -            & 90.46          & \cellcolor{blue!10}\textbf{82.16} & \textbf{77.31} \\
				\hline
			\end{tabular}
		\end{center}
		\caption{\small Performance comparison on KITTI \textit{test-set} with AP of 40 recall positions (AP40) for car class.}
		\label{tab:kitti_test}
	\end{table}
	
	\textbf{KITTI Dataset.}
	Firstly, we compare our 3D Cascade RCNN with several state-of-the-art 3D detectors on KITTI \textit{val-set}. Table \ref{tab:val_r11} and Table \ref{tab:kitti_val_ap40} summarize the performance comparisons in terms of AP11 and AP40 for three difficulties (\emph{i.e.}, Easy, Moderate, and Hard). Note that the metric of Moderate AP is officially adopted to rank the competing methods in the benchmark. The Easy AP refers to the evaluation over easy samples, which are fully visible with larger bounding box height and only accounts for a small part of all samples ($\sim$23.5\%). Hence the performance over Easy AP is almost saturated (\emph{e.g.}, SOTAs achieve $\sim$90\% in Table \ref{tab:val_r11}). The Hard AP additionally evaluates the completely difficult samples (\emph{e.g.}, the almost fully occluded samples) which are even difficult to see by humans. Instead, the Moderate AP performs a more comprehensive evaluation over a combination of easy samples and moderately difficult samples (\emph{e.g.}, the samples with relatively smaller bounding boxes or partly occluded samples). That’s why Moderate AP is officially adopted to rank the competing methods on KITTI Dataset. As such, the clear performance improvement over Moderate AP validates the effectiveness of our 3D Cascade RCNN over the easy and moderately difficult samples, since our re-weighting strategy encourages a more balanced loss across samples with different qualities. In general, under the major metric of Moderate AP, the results across both AP11 and AP40 consistently indicate that our 3D Cascade RCNN obtains better performances against existing techniques, including both one-stage detectors (\emph{e.g.}, SECOND, 3DSSD) and two-stage detectors (\emph{e.g.}, PV-RCNN, Voxel R-CNN). The results generally highlight key advantages of the cascade detection paradigm and completeness-aware re-weighting strategy for 3D object detection. Specifically, by leveraging a multi-level point-voxel integration strategy, PV-RCNN outperforms the detectors solely based on point inputs (\emph{e.g.}, PointPillars, PointRCNN) or voxel inputs (\emph{e.g.}, VoxelNet, SA-SSD). Next, Voxel R-CNN fully exploits the advantage of voxel representations through voxel RoI pooling, and thus boosts the performance. However, the Moderate AP11 and AP40 scores of Voxel R-CNN are still lower than our 3D Cascade RCNN, which not only strengthens the detection paradigm in a cascade manner but also targets mitigating sparsity problems via completeness-aware re-weighting.

	Similar observations are also attained in the performance comparison w.r.t Bird-Eye-View (BEV) AP, as summarized in Table \ref{tab:kitti_val_ap40}. In particular, the BEV AP40 on the Moderate level can achieve 92.41\%, making the absolute improvement over the best competitor Voxel R-CNN by 1.16\%.

	\begin{table}
		\small
		\begin{center}
			\setlength\tabcolsep{2pt}
			\begin{tabular}{c|cccc}
				\hline
				\multirow{2}{*}{Method}                       & \multicolumn{4}{c}{$\text{AP}_\text{3D}$ (\%) ~~}                                                      \\
				& Overall                                           & 0-30m          & 30-50m          & 50m-Inf         \\
				\hline
				\textbf{\textit{LEVEL\_1 3D mAP (IoU=0.7):}}  &                                                   &                &                 &                   \\
				PointPillar \cite{lang2019pointpillars}       & 56.62                                             & 81.01          & 51.75           & 27.94           \\
				MVF \cite{zhou2020end}                        & 62.93                                             & 86.30          & 60.02           & 36.02           \\
				Pillar-OD \cite{wang2020pillar}               & 69.80                                             & 88.53          & 66.50           & 42.93           \\
				AFDet+PointPillars-0.10 \cite{ge2020afdet}    & 63.69                                             & 87.38          & 62.19           & 29.27           \\
				PV-RCNN \cite{Shi_2020_CVPR}                  & 70.30                                             & 91.92          & 69.21           & 42.17           \\
				Voxel R-CNN \cite{deng2020voxel}              & 75.59                                             & 92.49          & 74.09           & 53.15           \\
				\textbf{3D Cascade RCNN}                      & \textbf{76.27}                                    & \textbf{92.66} & \textbf{74.99}  & \textbf{54.49}  \\
				\hline
				\textbf{\textit{LEVEL\_1 BEV mAP (IoU=0.7):}} &                                                   &                &                 &                   \\
				PointPillar \cite{lang2019pointpillars}       & 75.57                                             & 92.1           & 74.06           & 55.47           \\
				MVF \cite{zhou2020end}                        & 80.40                                             & 93.59          & 79.21           & 63.09           \\
				Pillar-OD \cite{wang2020pillar}               & 87.11                                             & 95.78          & 84.87           & 72.12           \\
				PV-RCNN \cite{Shi_2020_CVPR}                  & 82.96                                             & 97.35          & 82.99           & 64.97           \\
				Voxel R-CNN \cite{deng2020voxel}              & 88.19                                             & 97.62          & 87.34           & 77.70           \\
				\textbf{3D Cascade RCNN}                      & \textbf{88.25}                                    & \textbf{97.76} & \textbf{87.55}  & \textbf{78.03 } \\
				\hline
				\textbf{\textit{LEVEL\_2 3D mAP (IoU=0.7):}}  &                                                   &                &                 &                  \\
				PV-RCNN \cite{Shi_2020_CVPR}                  & 65.36                                             & 91.58          & 65.13           & 36.46           \\
				Voxel R-CNN  \cite{deng2020voxel}             & 66.59                                             & 91.74          & 67.89           & 40.80           \\
				\textbf{3D Cascade RCNN}                      & \textbf{67.12}                                    & \textbf{91.95} & \textbf{68.96}  & \textbf{41.82}  \\
				\hline
				\textbf{\textit{LEVEL\_2 BEV mAP (IoU=0.7):}} &                                                   &                &                                   \\
				PV-RCNN \cite{Shi_2020_CVPR}                  & 77.45                                             & 94.64          & 80.39           & 55.39           \\
				Voxel R-CNN \cite{deng2020voxel}              & 81.07                                             & 96.99          & 81.37           & 63.26           \\
				\textbf{3D Cascade RCNN}                      & \textbf{81.24}                                    & \textbf{97.16} & \textbf{81.65 } & \textbf{63.30}  \\
				\hline
			\end{tabular}
		\end{center}
		\caption{\small Performance comparison on Waymo Open Dataset for vehicle detection. AP is reported at different difficulty levels based on the object's distance from sensor (0-30m, 30-50m, 50m-inf) and the number of points within object (\textit{LEVEL\_1}, \textit{LEVEL\_2}).}
		\label{tab:waymo}
	\end{table}
	
	\begin{figure*}[!tb]
		\begin{center}
			\includegraphics[width=.245\textwidth]{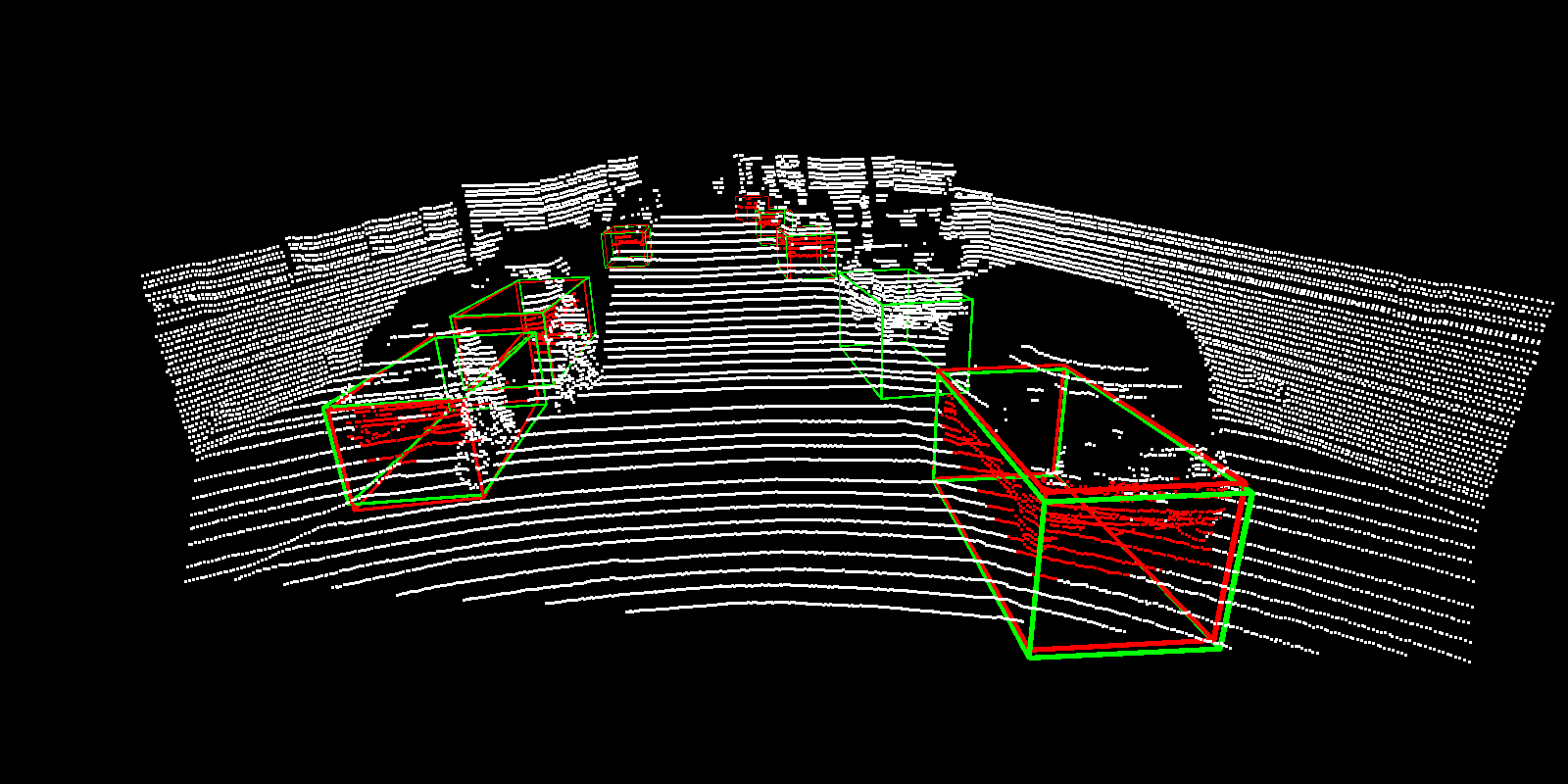}
			\includegraphics[width=.245\textwidth]{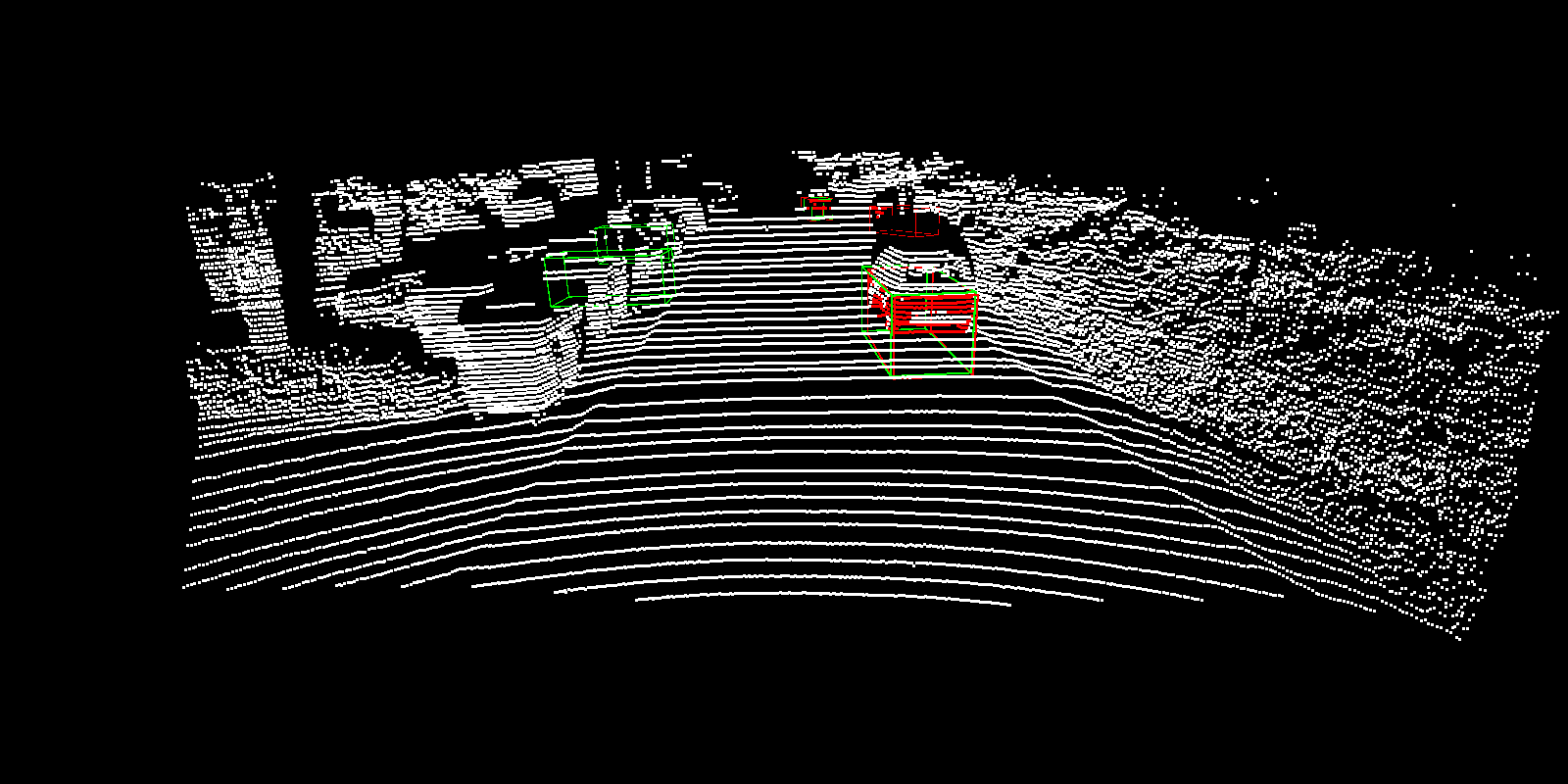}
			\includegraphics[width=.245\textwidth]{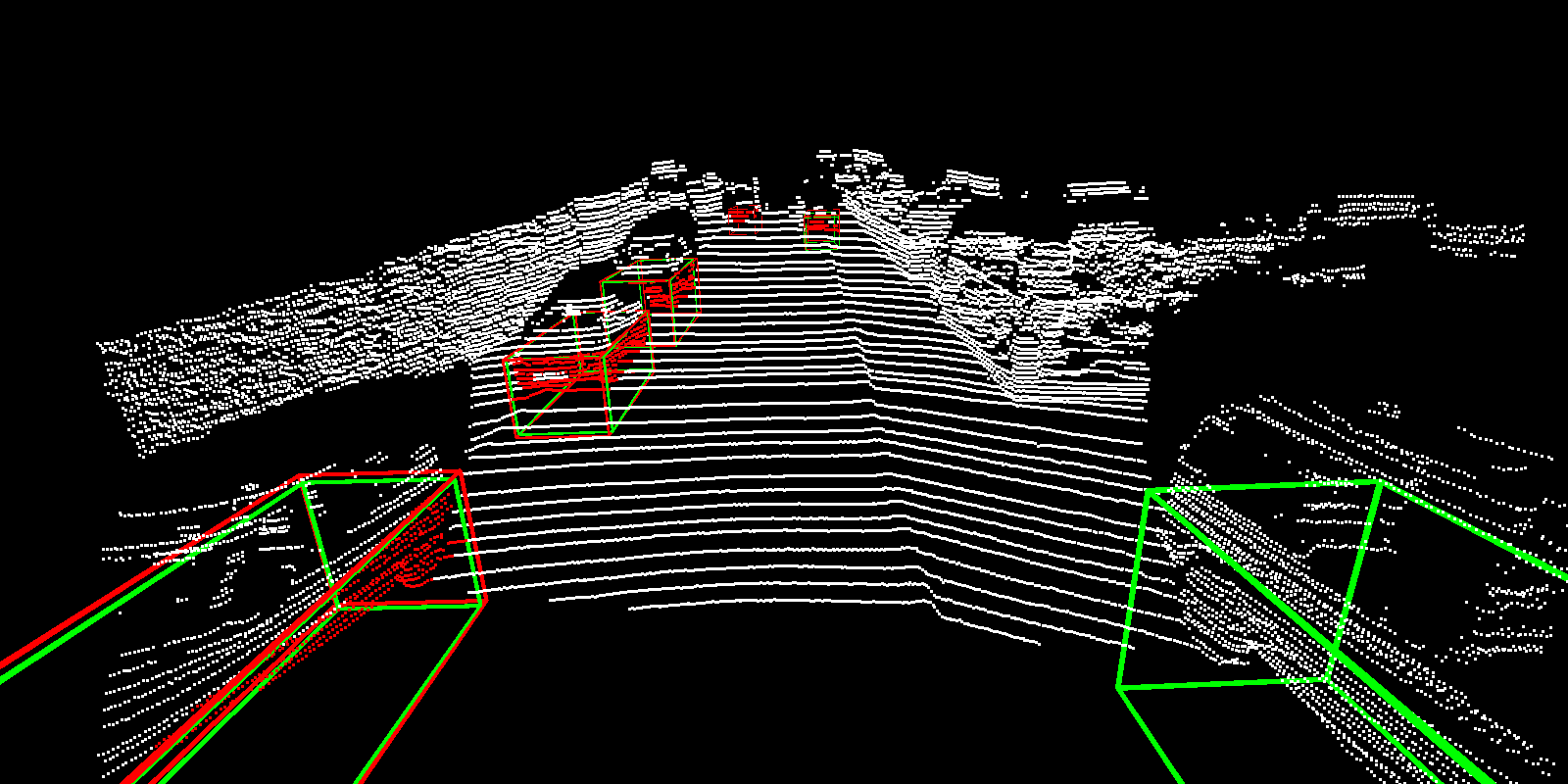}
			\includegraphics[width=.245\textwidth]{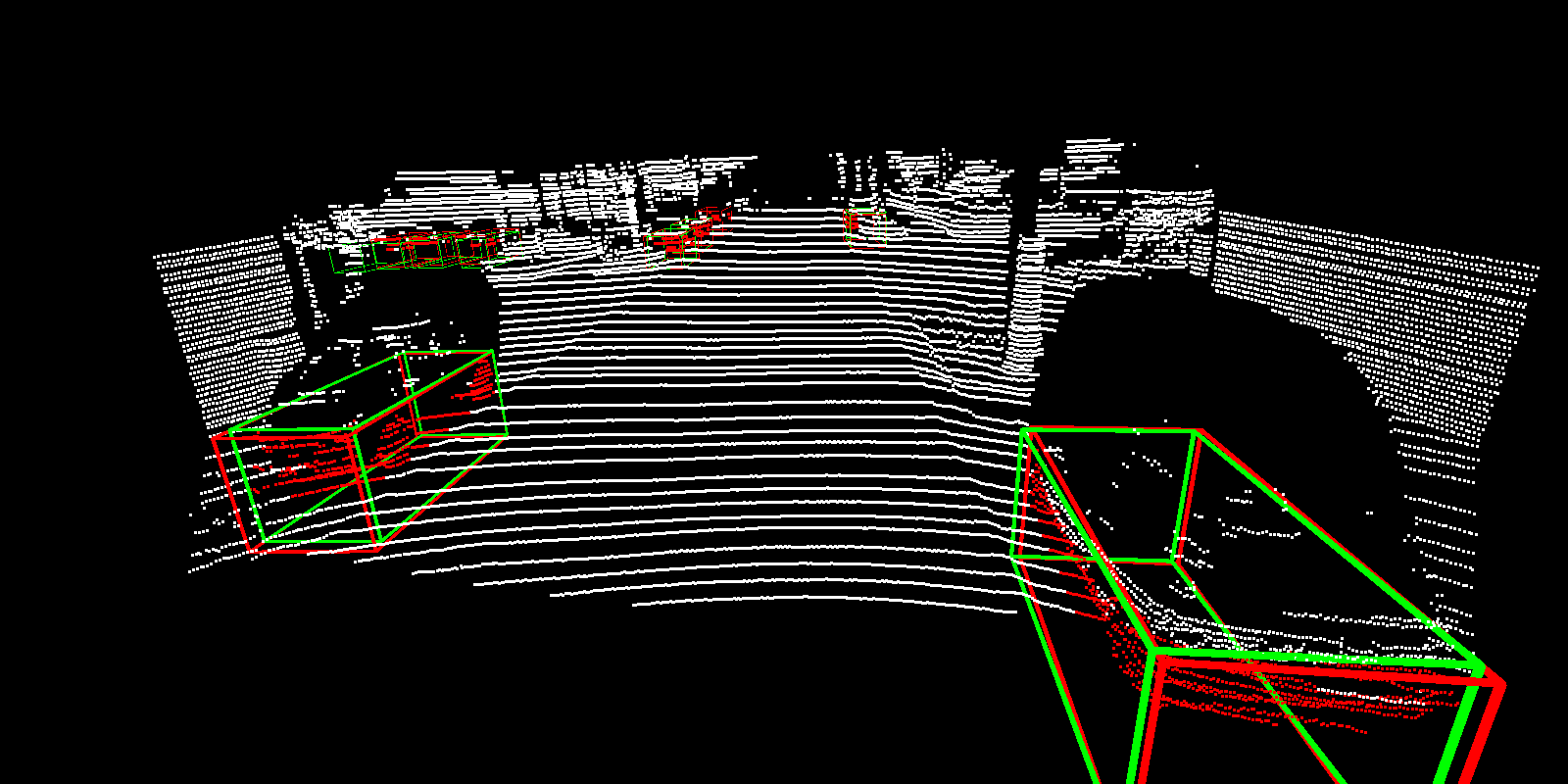}
			\includegraphics[width=.245\textwidth]{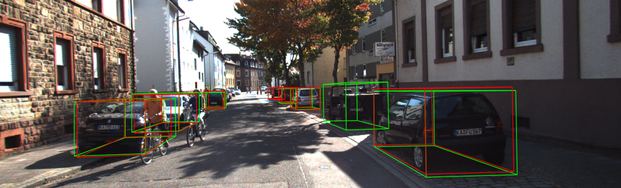}
			\includegraphics[width=.245\textwidth]{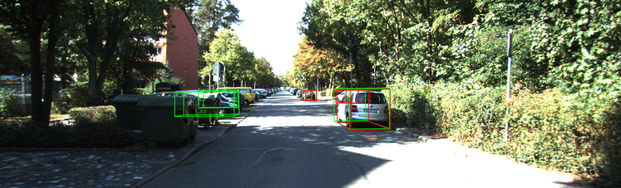}
			\includegraphics[width=.245\textwidth]{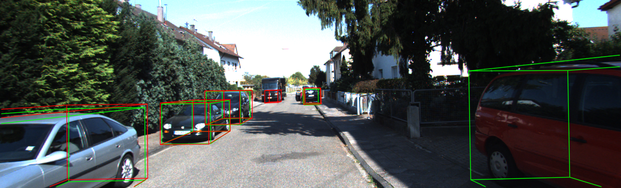}
			\includegraphics[width=.245\textwidth]{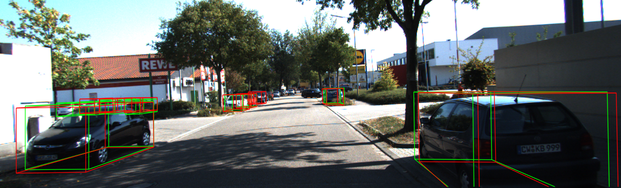}
		\end{center}
		\begin{center}
			\includegraphics[width=.245\textwidth]{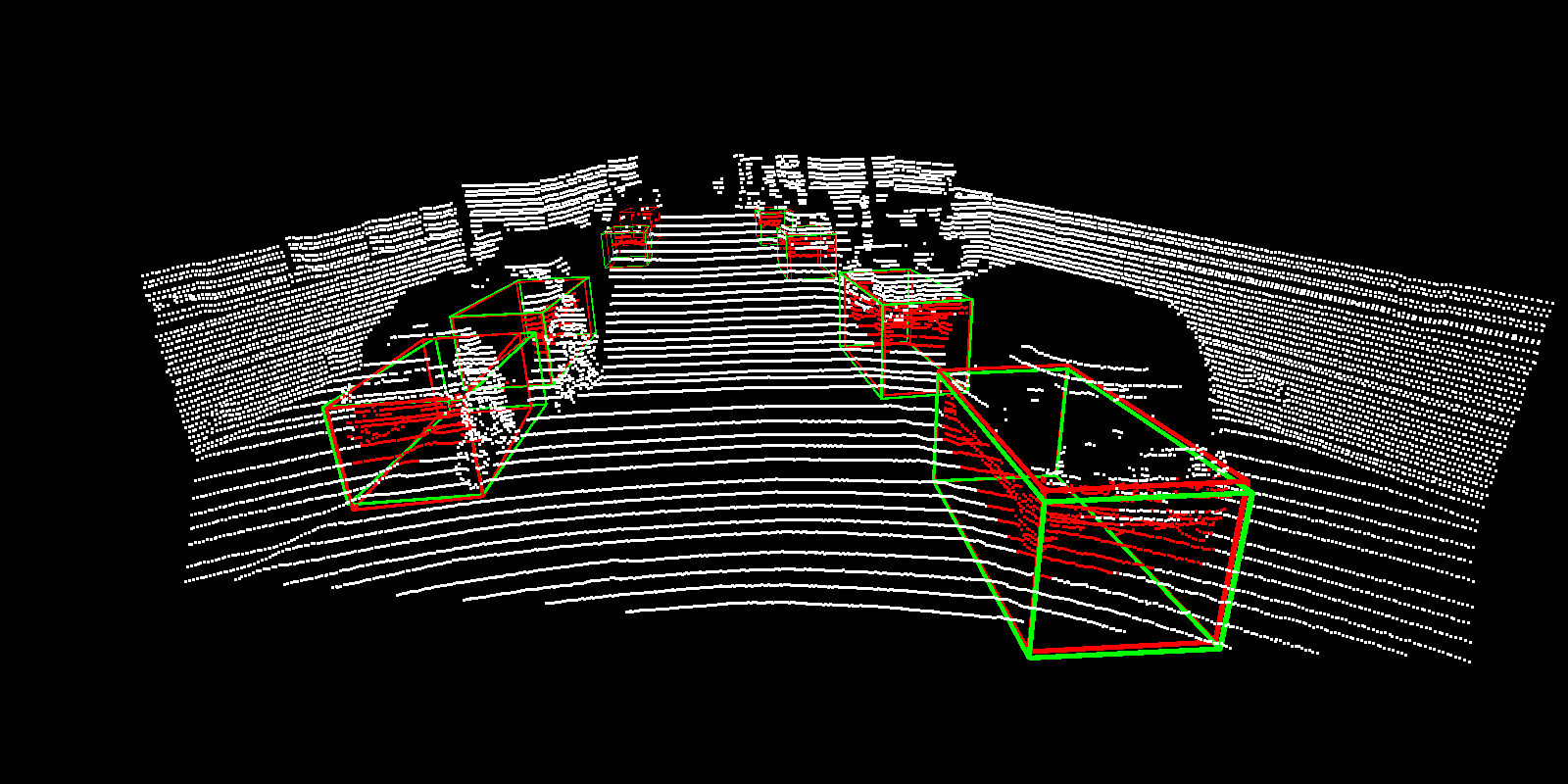}
			\includegraphics[width=.245\textwidth]{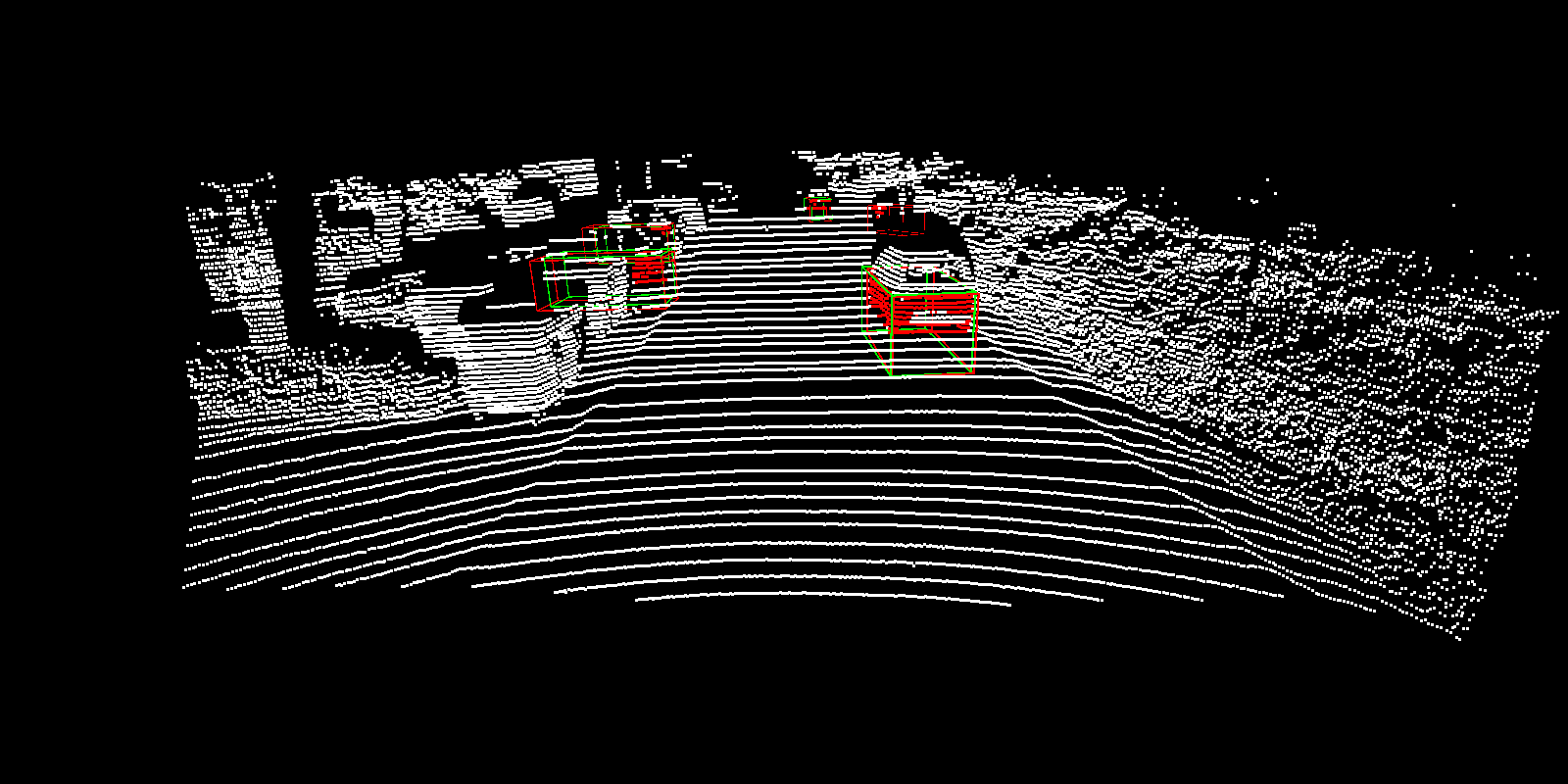}
			\includegraphics[width=.245\textwidth]{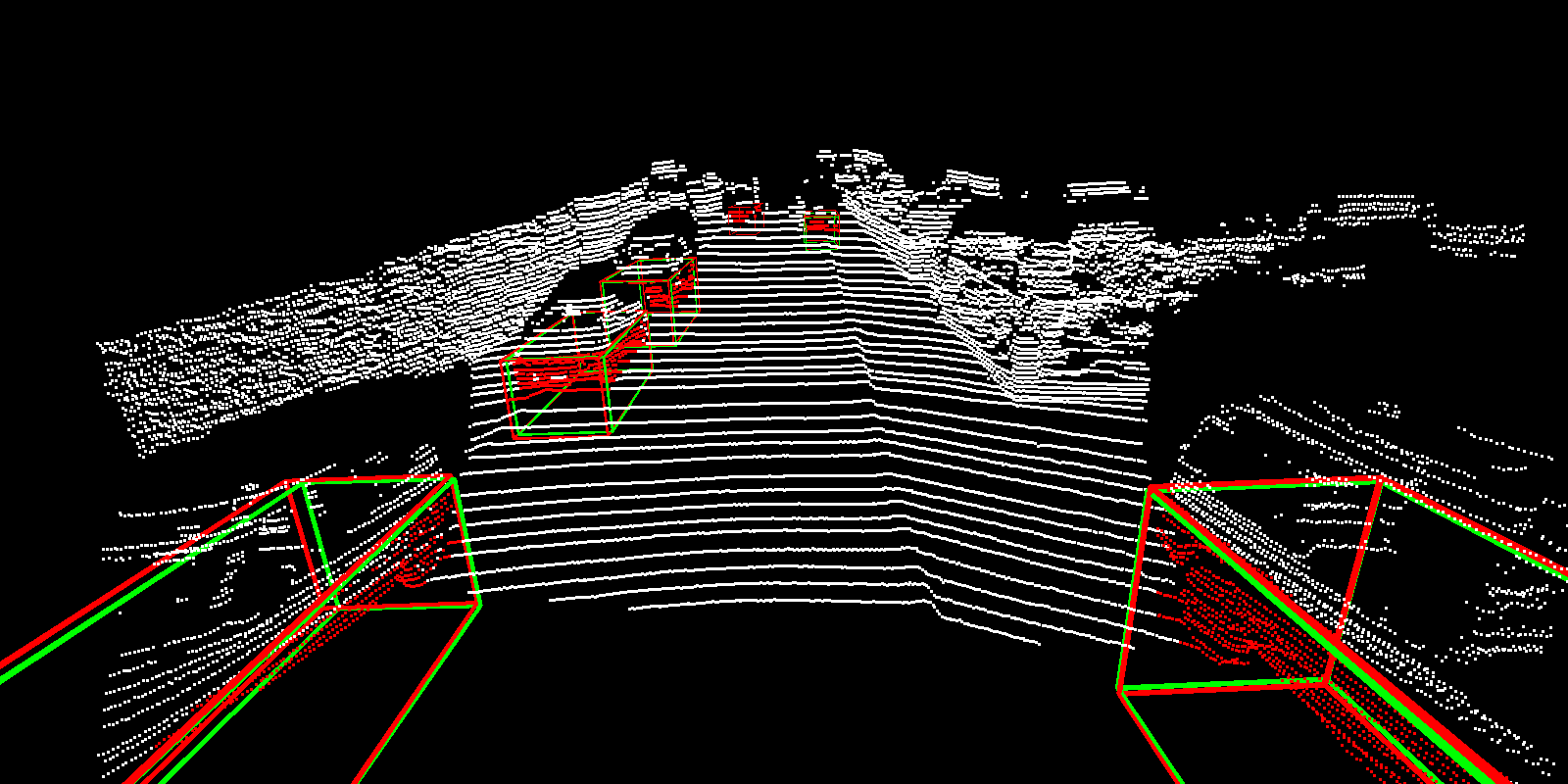}
			\includegraphics[width=.245\textwidth]{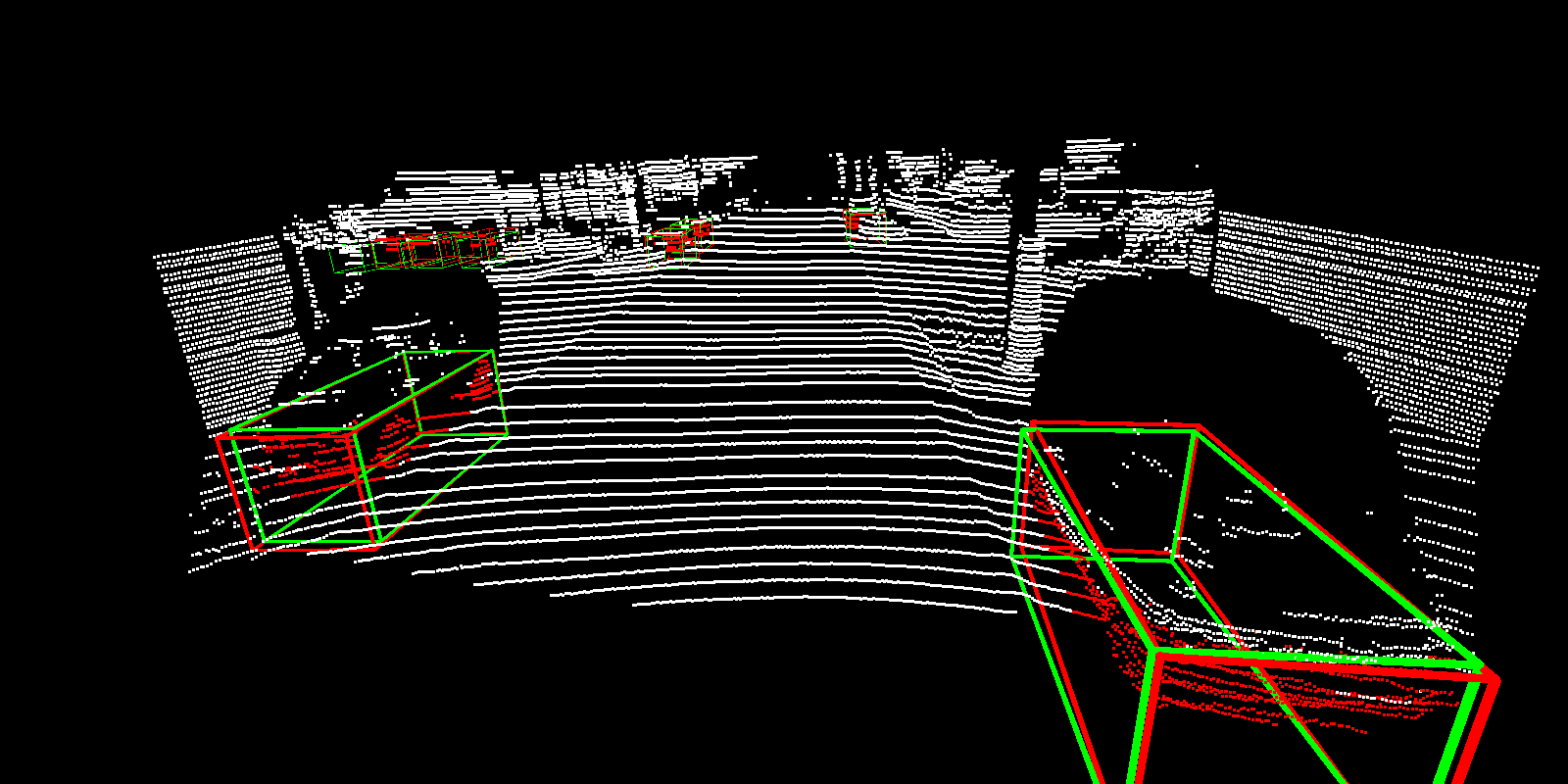}
			\includegraphics[width=.245\textwidth]{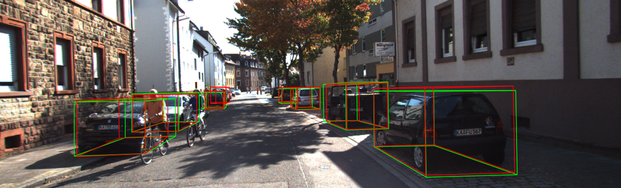}
			\includegraphics[width=.245\textwidth]{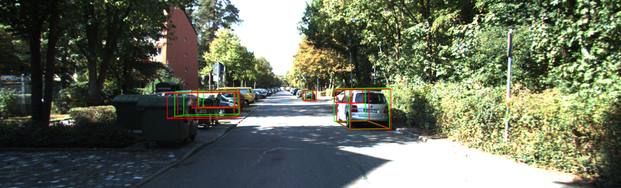}
			\includegraphics[width=.245\textwidth]{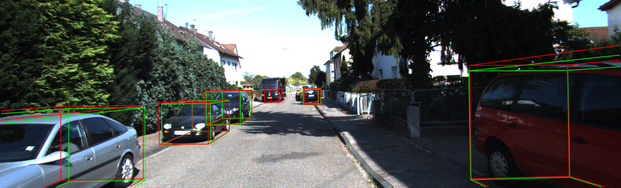}
			\includegraphics[width=.245\textwidth]{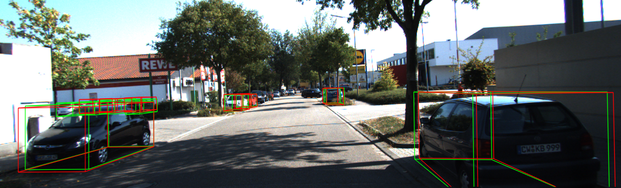}
		\end{center}
		\begin{center}
			\includegraphics[width=.245\textwidth]{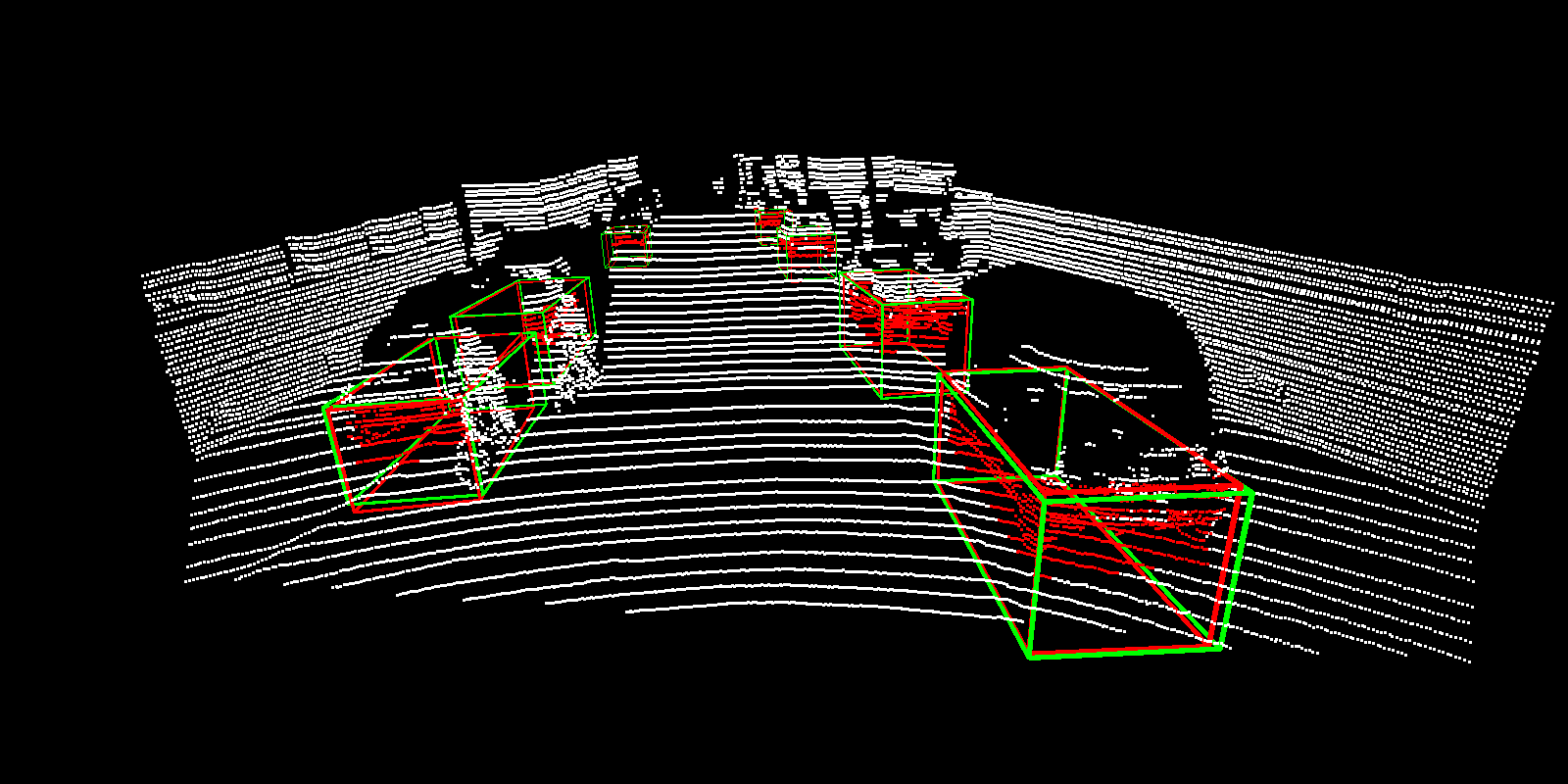}
			\includegraphics[width=.245\textwidth]{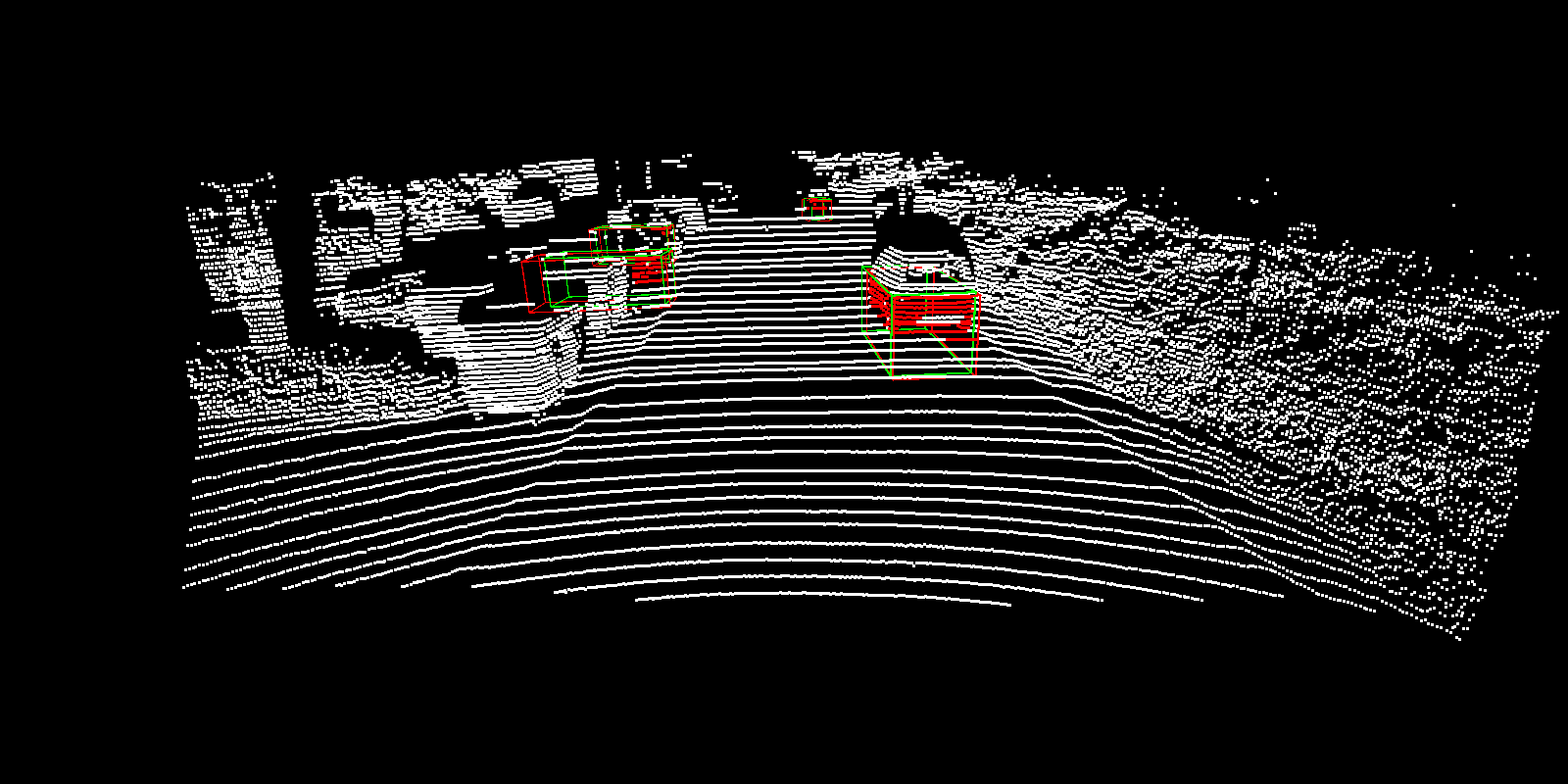}
			\includegraphics[width=.245\textwidth]{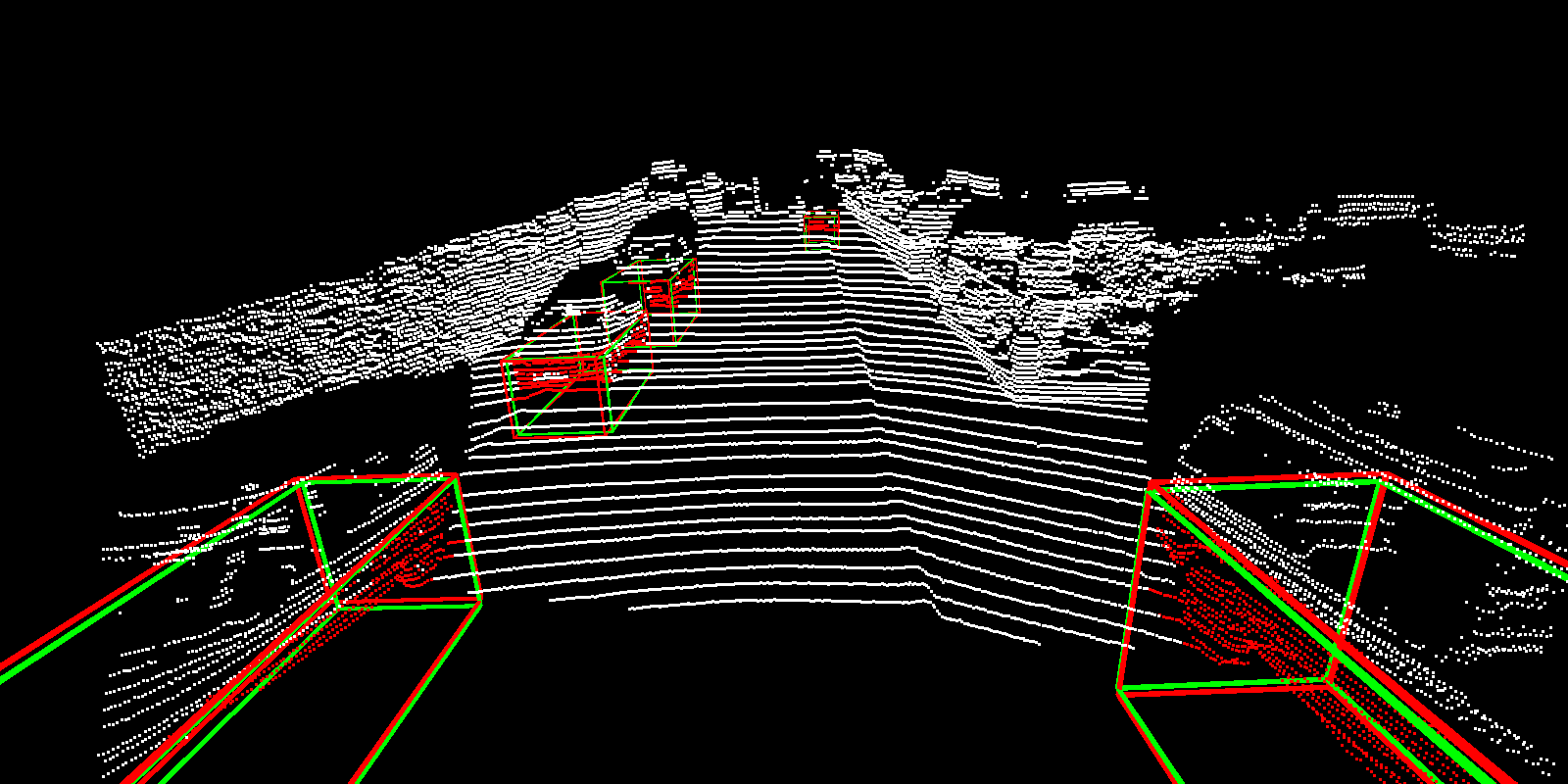}
			\includegraphics[width=.245\textwidth]{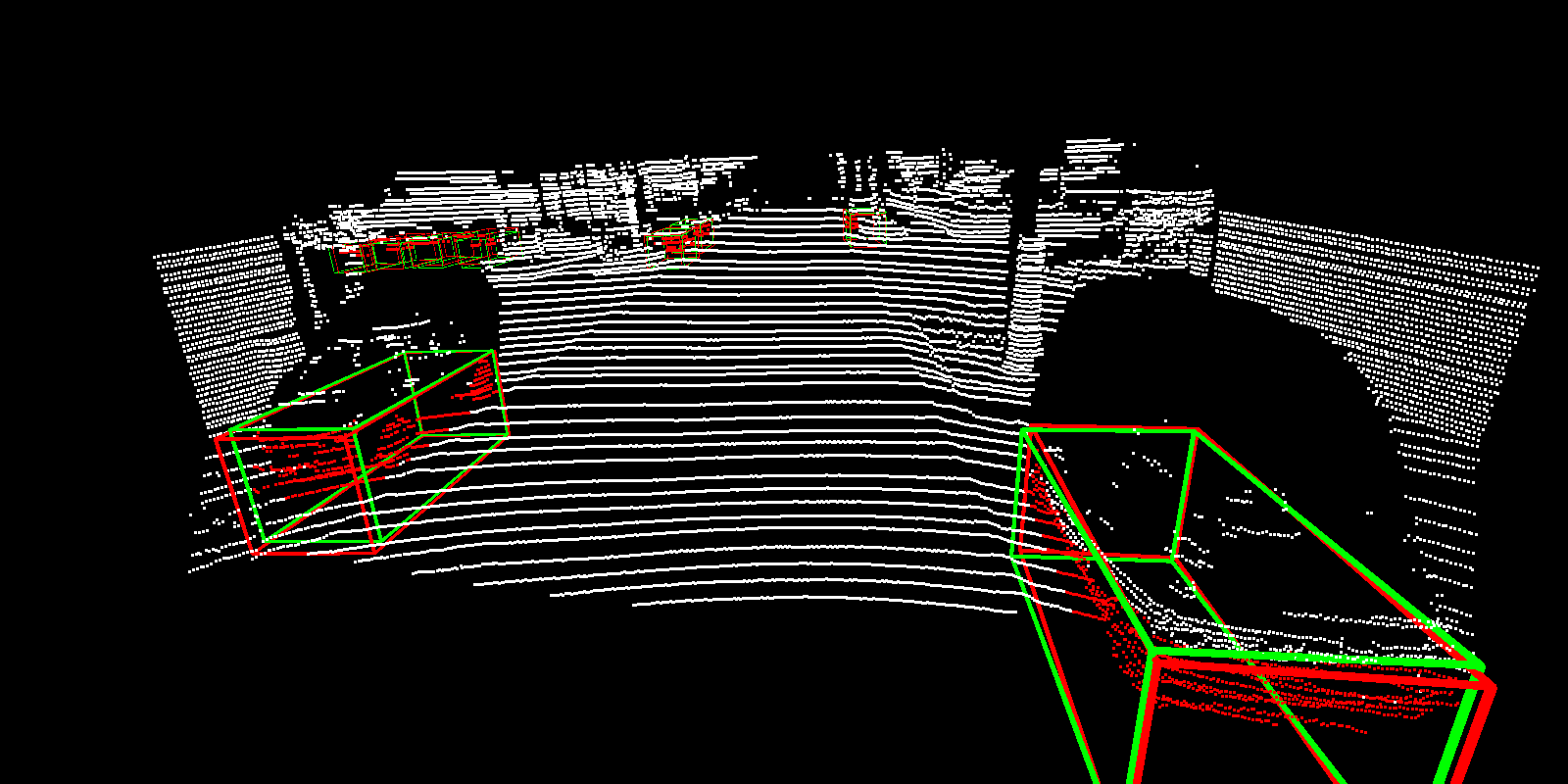}
			\includegraphics[width=.245\textwidth]{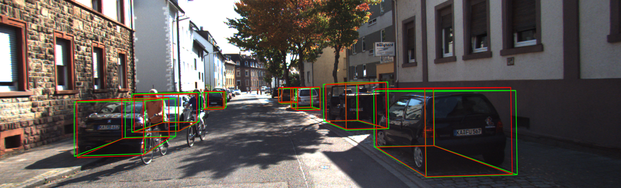}
			\includegraphics[width=.245\textwidth]{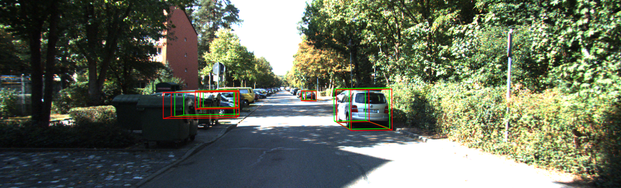}
			\includegraphics[width=.245\textwidth]{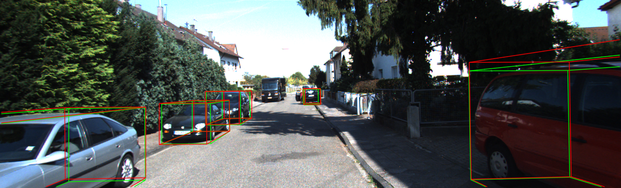}
			\includegraphics[width=.245\textwidth]{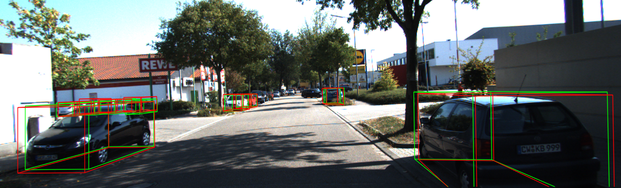}
		\end{center}
		\caption{\small Examples detection results of PV-RCNN (first row), Voxel R-CNN (second row) and our 3D Cascade RCNN (third row) on KITTI \textit{val-set}. For each method, the upper part is the view of the entire point clouds and  and the bottom part is the RGB image (Better viewed with zooming in). \textcolor{red}{RED} box denotes detection results and \textcolor{green}{GREEN} box is ground truth. Compared to the detection results of PV-RCNN and Voxel R-CNN, 3D Cascade RCNN generates fewer false alarm predictions (see the first two columns) and successfully detects remote cars with fewer points (see the last two columns).}
		\label{fig:example}
	\end{figure*}

	We also submitted our 3D Cascade RCNN to the official evaluation server and evaluated the performances for AP40 metric on KITTI \textit{test-set}. Table \ref{tab:kitti_test} shows the performances of 3D Cascade RCNN and state-of-the-art 3D detectors. Compared to the top-performing methods, our 3D Cascade RCNN manages to achieve the best AP40 score on the Moderate level. This again confirms that exploiting both cascade detection paradigm and completeness-aware re-weighting is an effective way for high-quality 3D object detection.

	Figure \ref{fig:example} showcases a few detection results generated by PV-RCNN, Voxel R-CNN and our 3D Cascade RCNN on KITTI \textit{test-set}. It is easy to see that both of the three detectors can recognize most of the target cars, while our 3D Cascade RCNN can predict more accurate 3D bounding boxes for the target cars as depicted in Figure \ref{fig:example}. Moreover,  the first two columns in Figure \ref{fig:example} show that our 3D Cascade RCNN generates fewer false alarms by suppressing their confidence scores. The results validate that compared to single-head detectors, the cascade detection paradigm with multi-stage regressors and confidence branches can lead to higher quality detections by iteratively refining the 3D proposals and confidence scores.

	\begin{figure*}
		\begin{center}
			\begin{minipage}{.3\textwidth}
				\animategraphics[autoplay,loop,width=1.0\textwidth]{8}{figs/waymo_results/segment-10247954040621004675_2180_000_2200_000/}{001}{018}
			\end{minipage}
			\begin{minipage}{.3\textwidth}
				\animategraphics[autoplay,loop,width=1.0\textwidth]{8}{figs/waymo_results/segment-12866817684252793621_480_000_500_000/}{001}{018}
			\end{minipage}
			\begin{minipage}{.3\textwidth}
				\animategraphics[autoplay,loop,width=1.0\textwidth]{8}{figs/waymo_results/segment-17694030326265859208_2340_000_2360_000/}{001}{018}
			\end{minipage}
		\end{center}
		\begin{center}
			\begin{minipage}{.3\textwidth}
				\animategraphics[autoplay,loop,width=1.0\textwidth]{8}{figs/waymo_results/segment-2506799708748258165_6455_000_6475_000/}{001}{018}
			\end{minipage}
			\begin{minipage}{.3\textwidth}
				\animategraphics[autoplay,loop,width=1.0\textwidth]{8}{figs/waymo_results/segment-5302885587058866068_320_000_340_000/}{001}{018}
			\end{minipage}
			\begin{minipage}{.3\textwidth}
				\animategraphics[autoplay,loop,width=1.0\textwidth]{8}{figs/waymo_results/segment-6491418762940479413_6520_000_6540_000/}{001}{018}
			\end{minipage}
		\end{center}
		\caption{\small Detection results of 3D Cascade RCNN on Waymo Open Dataset \emph{val} split (PLEASE view the animation in Acrobat Reader).}
		\label{fig:video_demo}
	\end{figure*}
	
	\textbf{Waymo Open Dataset.} We further evaluate our 3D Cascade RCNN on the more challenging Waymo Open Dataset. Table \ref{tab:waymo} shows the performance comparisons with state-of-the-art 3D detectors. Similar to the observations on KITTI, our 3D Cascade RCNN yields consistent gains against existing 3D detectors in terms of Overall AP$_\text{3D}$ and AP$_\text{BEV}$ at both \textit{LEVEL\_1} and \textit{LEVEL\_2}. This generally highlights the merit of iterative refinement of proposals and the additional guidance of point completeness score in our 3D Cascade RCNN for 3D object detection. Moreover, we present several video demos (GIF Animation) for the detection results of our 3D Cascade RCNN in Figure \ref{fig:video_demo}, which basically show the high quality detection of cars on Waymo Open Dataset.
	
	\subsection{Experimental Analysis}
	In this section, we provide more experimental analysis of our 3D Cascade RCNN. Ablation studies are first conducted to verify the effectiveness of each component in 3D Cascade RCNN, followed with the analysis with regard to the impact of stage number in cascade detection heads. Next, we present the performance comparisons among different IoU-based re-weighting strategies. Furthermore, a detailed experimental results of our 3D Cascade RCNN are shown over more classes (\emph{e.g.}, Car, Pedestrian and Cyclist) and different sparsity levels of objects. Finally, we evaluate the generalization capacity of our 3D Cascade RCNN by integrating existing state-of-the-art detectors (\emph{e.g.}, PV-RCNN) with our design and analyze different error types. Note that all the experiments here are conducted by training models on KITTI (Waymo) \textit{train-set} and evaluating them on KITTI (Waymo) \textit{val-set}.
	
	\subsubsection{Ablation Studies} \label{compare}
	To examine the impact of each design in 3D Cascade RCNN, we conduct ablation studies by comparing different variants of 3D Cascade RCNN in Table \ref{tab:component_ablation}. We start from the Base model (Model (a)), which is a basic two-stage detector (single detection head) without the completeness-aware re-weighting strategy. Next, we extend the Base model by utilizing three cascade detection heads that enable the iterative proposal refinement in a cascade manner, yielding Model (b) which achieves better performances. Meanwhile, by upgrading Model (a) with the completeness-aware re-weighting strategy, another variant of our model (Model (c)) also leads to performance boosts across all matrices. The performance improvements demonstrate the merit of tackling the point sparsity/quality problem in 3D object detection via completeness-aware re-weighting. Finally, the integration of cascade detection paradigm and completeness-aware re-weighting strategy (Model (d)), \emph{i.e.}, our 3D Cascade RCNN, obtains the highest performances across all the three difficulties, which validate the complementarity of the two designs.
	
	\begin{table}[!tb]
		\renewcommand\arraystretch{1}
		\small
		\begin{center}
			\setlength\tabcolsep{4pt}
			\begin{tabular}{c|cc|ccc}
				\hline
				\multicolumn{6}{l}{KITTI dataset \emph{val} split} \\ \hline
				\multirow{2}{*}{Methods} & Cascade      & \multirow{2}{*}{Re-weighting} & \multicolumn{3}{c}{$\text{AP}_\text{3D}$ (\%) ~~}                                                      \\
			                            & RoI Heads     &                               & Easy                                              & Moderate                          & Hard           \\
				\hline
				(a)                      &              &                               & 89.60                                             & \cellcolor{blue!10}84.47          & 78.84          \\
				(b)                      & $\bm{\surd}$ &                               & 89.68                                             & \cellcolor{blue!10}85.10          & 79.11          \\
				(c)                      &              & $\bm{\surd}$                  & 89.72                                             & \cellcolor{blue!10}85.53          & 79.03          \\
				(d)                      & $\bm{\surd}$ & $\bm{\surd}$                  & \textbf{90.05}                                    & \cellcolor{blue!10}\textbf{86.02} & \textbf{79.27} \\
				\hline
			\end{tabular}
		\begin{tabular}{c|cc|cc}
			\multicolumn{5}{l}{Waymo dataset \emph{val} split} \\ \hline
			\multirow{2}{*}{Methods} & Cascade      & \multirow{2}{*}{Re-weighting} & \multicolumn{2}{c}{$\text{AP}_\text{3D}$ (\%) ~~}                              \\
			
			& RoI Heads    &                               & \textbf{\textit{LEVEL\_1}}                        & \textbf{\textit{LEVEL\_2}} \\
			\hline
			(a)                      &              &                               & 75.59                                             & 66.59                      \\
			(b)                      & $\bm{\surd}$ &                               & 75.94                                             & 66.78                      \\
			(c)                      &              & $\bm{\surd}$                  & 75.83                                             & 66.70                      \\
			(d)                      & $\bm{\surd}$ & $\bm{\surd}$                  & \textbf{76.27}                                            & \textbf{67.12}                      \\
			\hline
		\end{tabular}
		\end{center}
		\caption{\small Ablation studies of 3D Cascade RCNN on KITTI \textit{val-set} and Waymo \textit{val-set}.}
		\label{tab:component_ablation}
	\end{table}
	
	\subsubsection{Impact of Stage Number in Cascade Detection Heads}
	Table \ref{tab:stage_num_ablation} shows the performances by varying the stage number in our 3D Cascade RCNN from 1 to 3. From observation, increasing the stage number can generally lead to performance improvement over AP11 at the Moderate level and the performances get saturated with 3 stages.
	
	\begin{table}[!tb]
		\renewcommand\arraystretch{1}
		\small
		\begin{center}
			\setlength\tabcolsep{17pt}
			\begin{tabular}{c|ccc}
				\hline
				\multirow{2}{*}{\#stages} & \multicolumn{3}{c}{$\text{AP}_\text{3D}$ (\%) ~~}                                                      \\
				& Easy                                              & Moderate                          & Hard           \\
				\hline
				1                         & 89.72                                             & \cellcolor{blue!10}85.53          & 79.03          \\
				2                         & 89.91                                             & \cellcolor{blue!10}85.95          & \textbf{79.34} \\
				3                         & \textbf{90.05}                                    & \cellcolor{blue!10}\textbf{86.02} & 79.27          \\
				\hline
			\end{tabular}
		\end{center}
		\caption{\small Impact of stage number in cascade detection heads on KITTI \textit{val-set} with AP of 11 recall positions (AP11).}
		\label{tab:stage_num_ablation}
	\end{table}
	
	\begin{table}[!tb]
		\renewcommand\arraystretch{1}
		\small
		\begin{center}
			\setlength\tabcolsep{15pt}
			\begin{tabular}{c|ccc}
				\hline
				\multirow{2}{*}{Re-Weighting} & \multicolumn{3}{c}{$\text{AP}_\text{3D}$ (\%) ~~}                                                      \\
				& Easy                                              & Moderate                          & Hard           \\
				\hline
				IoU-V1                        & 89.66                                             & \cellcolor{blue!10}85.11          & 79.12          \\
				IoU-V2                        & 89.57                                             & \cellcolor{blue!10}79.73          & 79.03          \\
				PC Score                      & \textbf{90.05}                                    & \cellcolor{blue!10}\textbf{86.02} & \textbf{79.27} \\
				\hline
			\end{tabular}
		\end{center}
		\caption{\small Performance comparisons of different re-weighting strategies on KITTI \textit{val-set} with AP of 11 recall positions (AP11).}
		\label{tab:iou_variant}
	\end{table}
	
	\begin{table*}[!tb]
		\renewcommand\arraystretch{1}
		\small
		\begin{center}
			\setlength\tabcolsep{10pt}
			\begin{tabular}{c|ccc|ccc|ccc}
				\hline
				\multirow{2}{*}{Classes}         & \multicolumn{3}{c}{$\text{Car AP}_\text{3D}$ (\%) ~~} & \multicolumn{3}{c}{$\text{Pedestrian AP}_\text{3D}$ (\%) ~~} & \multicolumn{3}{c}{$\text{Cyclist AP}_\text{3D}$ (\%) ~~}                                                                                                                                             \\
				& Easy                                                  & Moderate                                                     & Hard                                                      & Easy           & Moderate                          & Hard           & Easy           & Moderate                          & Hard           \\
				\hline
				PV-RCNN \cite{shi2020pv}         & 89.35                                                 & \cellcolor{blue!10}83.69                                     & 78.70                                                     & 63.12          & \cellcolor{blue!10}54.84          & 51.78          & 86.06          & \cellcolor{blue!10}69.48          & 64.50          \\
				Voxel R-CNN \cite{deng2020voxel} & 89.45                                                 & \cellcolor{blue!10}84.03                                     & 78.76                                                     & 66.74          & \cellcolor{blue!10}60.32          & 55.69          & 85.88          & \cellcolor{blue!10}71.53          & 68.20          \\
				3D Cascade RCNN                  & \textbf{90.05}                                        & \cellcolor{blue!10}\textbf{86.02}                            & \textbf{79.27}                                            & \textbf{69.32} & \cellcolor{blue!10}\textbf{61.71} & \textbf{57.48} & \textbf{87.04} & \cellcolor{blue!10}\textbf{73.91} & \textbf{72.29} \\
				\hline
			\end{tabular}
		\end{center}
		\caption{\small Performance comparisons over Car, Pedestrian and Cyclist classes on KITTI \textit{val-set} with AP of 11 recall positions (AP11).}
		\vspace{-0.2cm}
		\label{tab:more_classes}
	\end{table*}
	
	\begin{table}[!tb]
		\renewcommand\arraystretch{1}
		\small
		\begin{center}
			\setlength\tabcolsep{9pt}
			\begin{tabular}{c|ccc}
				\hline
				\multirow{2}{*}{Methods}         & \multicolumn{3}{c}{$\text{Moderate AP}_\text{3D}$ (\%) ~~}                                                      \\
				& Sparse                                                     & Modest                            & Complete       \\
				\hline
				PV-RCNN \cite{shi2020pv}         & 14.50                                                      & \cellcolor{blue!10} 37.70         & 93.27          \\
				Voxel R-CNN \cite{deng2020voxel} & 16.28                                                      & \cellcolor{blue!10}42.03          & {93.81}        \\
				3D Cascade RCNN                  & \textbf{17.14}                                             & \cellcolor{blue!10}\textbf{44.60} & \textbf{94.62} \\
				\hline
			\end{tabular}
		\end{center}
		\caption{\small Performance comparisons of different sparsity levels on KITTI \textit{val-set} with AP of 11 recall positions (AP11).}
		\label{tab:sparsity_aware}
	\end{table}
	
	\begin{table}[!tb]
		\renewcommand\arraystretch{1}
		\small
		\begin{center}
			\setlength\tabcolsep{8pt}
			\begin{tabular}{c|ccc}
				\hline
				\multirow{2}{*}{Methods} & \multicolumn{3}{c}{$\text{AP}_\text{3D}$ (\%) ~~}                                                      \\
				& Easy                                              & Moderate                          & Hard           \\
				\hline
				PV-RCNN                  & 89.35                                             & \cellcolor{blue!10}83.69          & 78.70          \\
				PV-RCNN w/ 3D Cascade    & \textbf{89.37}                                    & \cellcolor{blue!10}\textbf{84.33} & \textbf{78.96} \\
				\hline
			\end{tabular}
		\end{center}
		\caption{\small Performance comparisons by integrating PV-RCNN with the designs in our 3D Cascade RCNN on KITTI \textit{val-set} with AP of 11 recall positions (AP11).
		}
		\label{tab:pvrcnn_with_cascade}
	\end{table}

	\subsubsection{Comparisons with IoU Based Re-weighting} \label{sec:iou}
	To further understand the effects of completeness-aware re-weighting, we design two variants of IoU based re-weighting strategies. Specifically, we first utilize the IoU between the proposal and ground truth box to replace the PC score in Eq. (\ref{eq:linear_task_weight}). This variant is named as IoU-V1. For another variant (IoU-V2), we directly replace the PC score with the IoU between the proposal and the smallest enclosing box of points. As shown in Table \ref{tab:iou_variant}, our adopted PC Score based re-weighting strategy achieves better performances against the other two IoU-based variants. The results generally demonstrate the advantage of our proposed Point Completeness Score module for sample re-weighting in 3D Cascade RCNN.

	\subsubsection{Performances on More Classes}
	In Table \ref{tab:more_classes}, we present the performance comparisons over three classes, \emph{i.e.}, Car, Pedestrian and Cyclist, on KITTI \emph{val} split. Following PV-RCNN \cite{shi2020pv}, we set the IoU threshold as 0.7 for Car, and 0.5 for Pedestrian and Cyclist classes. As shown in Table \ref{tab:more_classes}, our 3D Cascade RCNN achieves 1.4\%+ performance boost against PV-RCNN \cite{shi2020pv} and Voxel R-CNN \cite{deng2020voxel} on Moderate AP metric across all the three classes. This generally corroborate the generalization of our 3D Cascade RCNN when applied for 3D object detection over different object categories.
	
	\subsubsection{Performance Improvements on Different Sparsity Levels}
	Recall that our devised completeness-aware re-weighting strategy down-weights the contributions of sparse objects during training. Hence a natural question is whether such strategy will decline the performances on sparse objects. To explore this point, we additionally report the performances of our 3D Cascade RCNN on KITTI \emph{val} with different sparsity levels. Note that we group ground truth objects into three levels according to their PC scores: Sparse (PC Score $<$ 0.3), Modest (PC Score $\ge$ 0.3 and PC Score $<$ 0.6), and Complete (PC Score $\ge$ 0.6). As illustrated in Table \ref{tab:sparsity_aware}, our 3D Cascade RCNN exhibits better performances than PV-RCNN and Voxel R-CNN across all sparsity levels, which validate the merit of our completeness-aware re-weighting by balancing the losses of samples in all sparsity levels.

\begin{figure}[!tb]
		\centering {\includegraphics[width=0.46\textwidth]{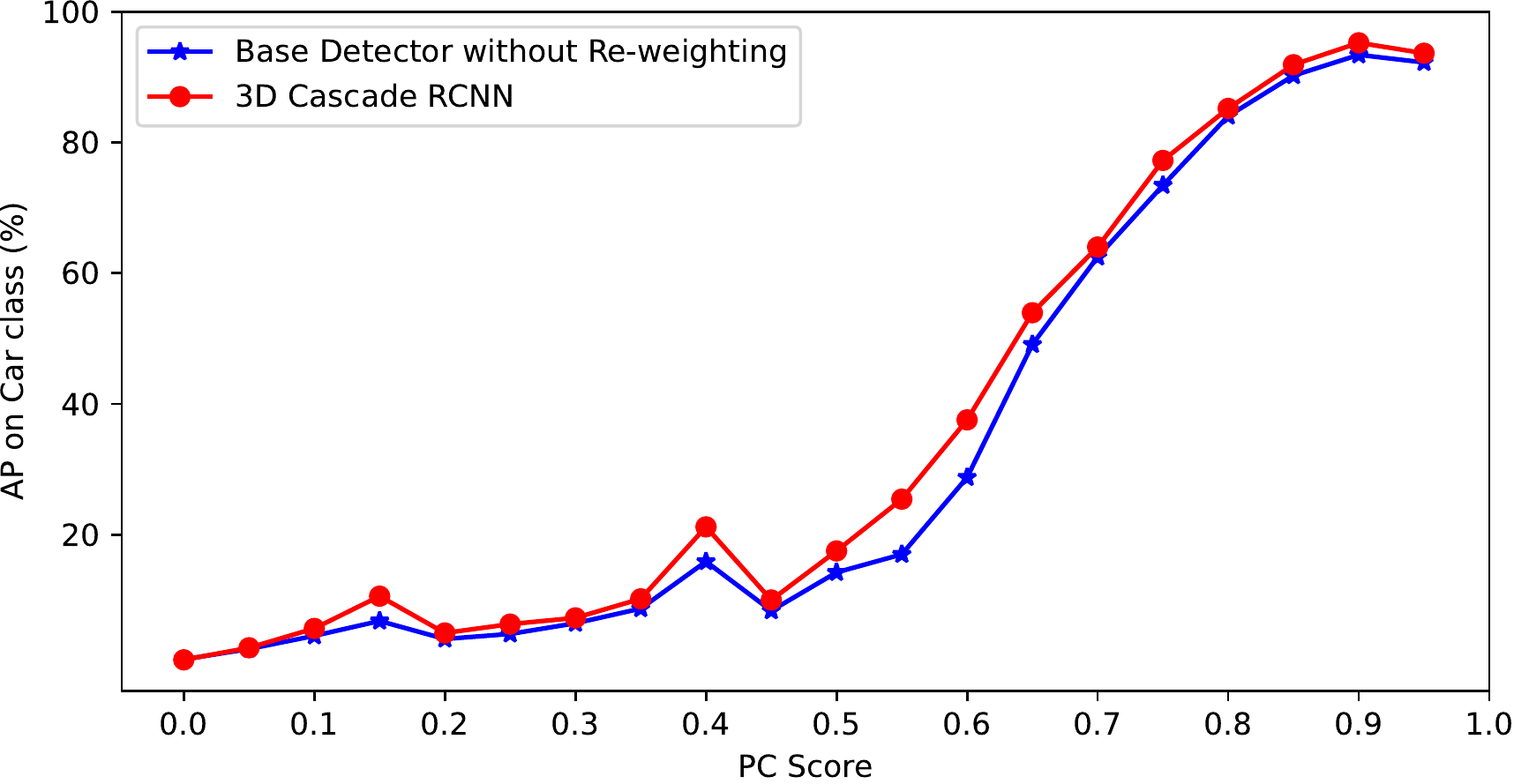}}
		\caption{\small Performance distributions under different PC Score intervals on KITTI \emph{val-set} with AP of 11 recall positions (AP11) for car class.}
		\label{fig:ap_distribution}
	\end{figure}

	As suggested, we additionally visualize the performance distributions for our 3D Cascade RCNN and the Base detector (without Re-weighting) under different PC Score intervals. As shown in Fig. \ref{fig:ap_distribution}, 3D Cascade RCNN consistently achieves better performances than Base detector under most PC Score intervals. In particular, clear performance improvements are attained when PC Score varies within the range from 0.3 to 0.8. For very sparse objects (\emph{i.e.}, the almost fully occluded samples with PC Score $<$ 0.3) that are even difficult to identify by humans, 3D Cascade RCNN still manages to outperform Base detector. For easy objects with PC Score $>$ 0.8 which are commonly visible with larger bounding box height, the performance is almost saturated and it is relatively difficult to introduce large margin of improvement.

	\subsubsection{Evaluation of Generalization Capacity}
	To fully verify the generalizability of the designs in our 3D Cascade RCNN for 3D object detection, we evaluate the performances of PV-RCNN \cite{shi2020pv} by integrating it with our devised cascade detection paradigm plus completeness-aware re-weighting strategy. Specifically, we upgrade the single detection head in PV-RCNN with two cascade detection heads and additionally leverage the point completeness score module for completeness-aware re-weighting. As shown in Table \ref{tab:pvrcnn_with_cascade}, our designs consistently boost up the Moderate AP11 performances, which again confirms the generalizability capacity of our proposals.
	
	\begin{figure}[!tb]\label{fig:intro_demo}
		\centering {\includegraphics[width=0.5\textwidth]{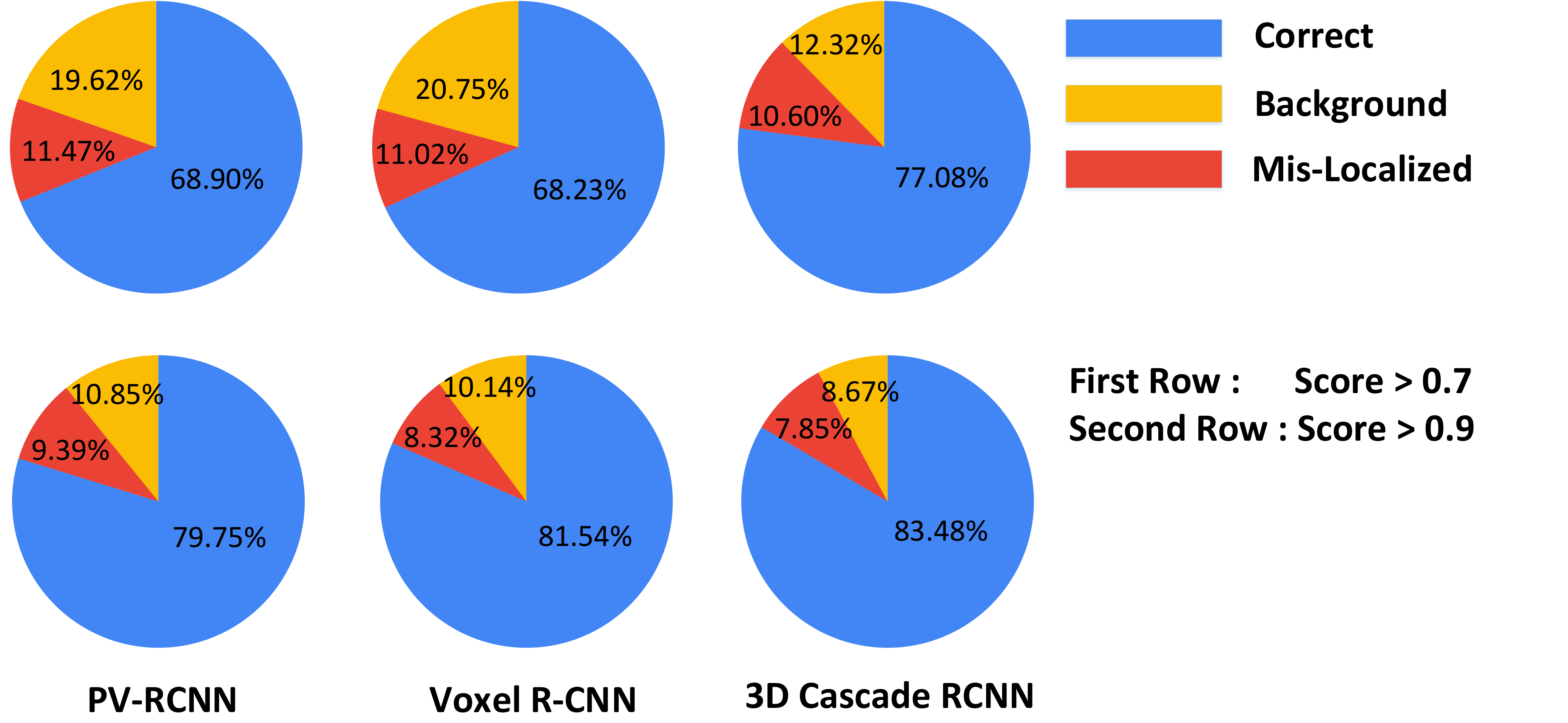}}
		\caption{\small Error analysis of high-confident detections on KITTI \emph{val} split. The detection errors are categorized into two types: poor localization (Mis-Localized) and false alarms (Background). The first row shows detections with confidence score $>$ 0.7 and the second row depicts the detection with confidence score $>$ 0.9.}
		\label{fig:error_types}
	\end{figure}
	
	Since we design PC Score-based re-weighting to be a unified re-weighting strategy that re-weight the task weight of each proposal according to its point completeness score, it is feasible to plug our PC Score-based re-weighting into any single-stage/two-stage detectors. We additionally incorporate PC Score-based re-weighting into single-stage detectors (\emph{e.g.}, PointPillars and SECOND) under the codebase of OpenPCDet. As shown in Table \ref{tab:one_stage_with_r}, our PC Score-based re-weighting leads to performance boosts for both PointPillars and SECOND, which basically validate the generalizability capacity of this re-weighting strategy.

	\begin{table}[!tb]
		\renewcommand\arraystretch{1}
		\small
		\begin{center}
			\setlength\tabcolsep{8pt}
			\begin{tabular}{c|ccc}
				\hline
				\multirow{2}{*}{Methods}                 & \multicolumn{3}{c}{$\text{AP}_\text{3D}$ (\%) ~~}                                                      \\
				& Easy                                              & Moderate                          & Hard           \\
				\hline
				PointPillars \cite{lang2019pointpillars} & 86.30                                             & \cellcolor{blue!10}75.54          & 72.45          \\
				PointPillars + Re-weighting              & \textbf{87.05}                                    & \cellcolor{blue!10}\textbf{76.21} & \textbf{73.32} \\
				
				SECOND \cite{yan2018second}              & 88.94                                             & \cellcolor{blue!10}79.19          & 75.88          \\
				SECOND + Re-weighting                    & \textbf{89.15}                                    & \cellcolor{blue!10}\textbf{79.93} & \textbf{76.76} \\
				\hline
			\end{tabular}
		\end{center}
		\caption{\small Performance comparisons by integrating PC Score-based re-weighting into PointPillars and SECOND on KITTI \textit{val-set} with AP of 11 recall positions (AP11).
		}
		\label{tab:one_stage_with_r}
	\end{table}

	\subsubsection{Error Analysis of High-Confident Detections}
	Here we further analyze the detection errors of our 3D Cascade RCNN and existing techniques (\emph{e.g.}, PV-RCNN \cite{shi2020pv} and Voxel R-CNN \cite{deng2020voxel}) on KITTI \emph{val} split. In particular, we categorize the detection results into three types: \textbf{Correct} (IoU with ground truth $\in [0.7, 1.0]$), \textbf{Mis-Localized} (IoU with ground truth $\in [0.5, 0.7)$), and \textbf{Background} (IoU with ground truth $\in [0.0, 0.5)$). We firstly filter less-confident detections whose confidence scores are lower than a threshold (0.7 and 0.9 respectively). Then we compute the IoUs of retained high-confident detections with ground truth boxes, and finally report the ratios of each detection type in Figure \ref{fig:error_types}. Compared to PV-RCNN and Voxel R-CNN, our 3D Cascade RCNN clearly reduces the number of each kind of detection error (\emph{i.e.}, Mis-Localized and Background). Most specifically, under the score threshold of 0.7, the Background error is significantly decreased by 40\%, which shows that our 3D Cascade RCNN can avoid more false alarms. When increasing the confidence threshold (from 0.7 to 0.9), the ratio of each error type decreases for each method.

	\subsubsection{IoU Thresholds at Different Stages}
	In the design of 3D Cascade RCNN, we use fixed-IoU threshold among different stages, which is different from 2D Cascade R-CNN where the IoU thresholds for matching positive samples are increased as the stage going deeper. We adopt the fixed-IoU threshold for two main reasons. Most importantly, 2D cascade RCNN is evaluated over MSCOCO dataset where the mmAP metric (\emph{i.e.}, averaged AP across different IoU thresholds) is adopted for COCO-style evaluation. On the contrary, 3D object detection techniques utilize AP for evaluation, which is measured under only one fixed IoU threshold. That’s why we use the fixed-IoU threshold for each stage, which is better aligned with the fixed-IoU evaluation metric in 3D object detection. Moreover, the using of fixed-IoU threshold significantly reduces the cost of hyperparameter tuning at each stage. It is worthy to note that we indeed experimented by increasing IoU thresholds (\emph{i.e.}, [0.75, 0.85, 0.95]) in the first, second and third stages. As shown in the Table \ref{tab:iou_inccrease}, increasing IoU thresholds stage-by-stage will result in the performance drop over Moderate AP, which validates the effectiveness of our fixed-IoU threshold at each stage.

	\begin{table}[!tb]
		\renewcommand\arraystretch{1}
		\small
		\begin{center}
			\setlength\tabcolsep{8pt}
			\begin{tabular}{c|ccc}
				\hline
				\multirow{2}{*}{IoU Thresholds} & \multicolumn{3}{c}{$\text{AP}_\text{3D}$ (\%) ~~}                                                      \\
				& Easy                                              & Moderate                          & Hard           \\
				\hline
				$[0.75, 0.85, 0.95]$              & 89.82                                             & \cellcolor{blue!10}85.80          & 79.27          \\
				$[0.75, 0.75, 0.75]$              & \textbf{90.05}                                    & \cellcolor{blue!10}\textbf{86.02} & \textbf{79.27} \\
				\hline
			\end{tabular}
		\end{center}
		\caption{\small Performance comparisons by different IoU thresholds settings on KITTI \emph{val} split with AP of 11 recall positions (AP11)..
		}
		\label{tab:iou_inccrease}
	\end{table}
	
	\subsubsection{Run Time Comparison}
	We further examine the performance tradeoff of 3D Cascade RCNN. We include the detailed run time and performance of each method in Table \ref{tab:speed}. The results show that our 3D Cascade RCNN achieves comparable run time against state-of-the-art single-head approach (Voxel R-CNN). Note that although our 3D Cascade RCNN has multiple cascade heads that result in additional computational cost, we slim the channels of RPN and RoI heads, thereby pursuing a better cost-performance tradeoff.

	\begin{table}[!tb]
		\renewcommand\arraystretch{1}
		\small
		\begin{center}
			\setlength\tabcolsep{8pt}
			\begin{tabular}{c|ccc|c}
				\hline
				\multirow{2}{*}{Methods} & \multicolumn{3}{c}{$\text{AP}_\text{3D}$ (\%) ~~} & \multirow{2}{*}{FPS}                 \\
				& Easy                                              & Moderate             & Hard  &       \\
				\hline
				PV-RCNN                  & 89.35                                             & 83.69                & 78.70 & 5.96  \\
				Voxel R-CNN              & 89.41                                             & 84.52                & 78.93 & 17.17 \\
				3D Cascade RCNN                  & \textbf{90.05}                                            & \textbf{86.02 }               &\textbf{ 79.27} & 14.16 \\
				\hline
			\end{tabular}
		\end{center}
		\caption{\small Performance and run-time comparisons of 3D Cascade RCNN and base detectors on KITTI \emph{val-set}.
		}
		\label{tab:speed}
	\end{table}
	
	\section{Conclusions}
	In this work, we propose a simple, yet high-quality detector, termed as 3D Cascade RCNN, by remolding the generic cascade detection paradigm in 2D images for 3D object detection. Particularly, we study the problem from the viewpoint of delving into the inherent sparsity problem caused by highly sparse input data under the cascade paradigm. To verify our claim, we quantitatively define the sparsity level of the observed points for each object as the point completeness score. The point completeness score of each proposal is further leveraged as the task weight for training the detector at each stage. Such completeness-aware re-weighting strategy targets for screening out the high-quality proposals with relatively complete point distribution and thus enables a more stable training process. Experiments conducted on KITTI and Waymo Open Dataset demonstrate our proposal and analysis. We also demonstrate the generalization of our 3D Cascade RCNN by integrating other 3D detectors (\emph{e.g.}, PV-RCNN) with our cascade paradigm plus completeness-aware re-weighting strategy.
	
	{
		\bibliographystyle{IEEEtran}
		\bibliography{egbib}

\begin{thebibliography}{10}
\providecommand{\url}[1]{#1}
\csname url@samestyle\endcsname
\providecommand{\newblock}{\relax}
\providecommand{\bibinfo}[2]{#2}
\providecommand{\BIBentrySTDinterwordspacing}{\spaceskip=0pt\relax}
\providecommand{\BIBentryALTinterwordstretchfactor}{4}
\providecommand{\BIBentryALTinterwordspacing}{\spaceskip=\fontdimen2\font plus
\BIBentryALTinterwordstretchfactor\fontdimen3\font minus
  \fontdimen4\font\relax}
\providecommand{\BIBforeignlanguage}[2]{{%
\expandafter\ifx\csname l@#1\endcsname\relax
\typeout{** WARNING: IEEEtran.bst: No hyphenation pattern has been}%
\typeout{** loaded for the language `#1'. Using the pattern for}%
\typeout{** the default language instead.}%
\else
\language=\csname l@#1\endcsname
\fi
#2}}
\providecommand{\BIBdecl}{\relax}
\BIBdecl

\bibitem{qi2017pointnet}
C.~R. Qi, H.~Su, K.~Mo, and L.~J. Guibas, ``Pointnet: Deep learning on point
  sets for 3d classification and segmentation,'' in \emph{Proceedings of the
  IEEE conference on computer vision and pattern recognition}, 2017, pp.
  652--660.

\bibitem{shi2019pointrcnn}
S.~Shi, X.~Wang, and H.~Li, ``Pointrcnn: 3d object proposal generation and
  detection from point cloud,'' in \emph{Proceedings of the IEEE conference on
  computer vision and pattern recognition}, 2019, pp. 770--779.

\bibitem{yan2018second}
Y.~Yan, Y.~Mao, and B.~Li, ``Second: Sparsely embedded convolutional
  detection,'' \emph{Sensors}, p. 3337, 2018.

\bibitem{zhou2018voxelnet}
Y.~Zhou and O.~Tuzel, ``Voxelnet: End-to-end learning for point cloud based 3d
  object detection,'' in \emph{Proceedings of the IEEE conference on computer
  vision and pattern recognition}, 2018, pp. 4490--4499.

\bibitem{chen2017multi}
X.~Chen, H.~Ma, J.~Wan, B.~Li, and T.~Xia, ``Multi-view 3d object detection
  network for autonomous driving,'' in \emph{Proceedings of the IEEE conference
  on computer vision and pattern recognition}, 2017, pp. 1907--1915.

\bibitem{ku2018joint}
J.~Ku, M.~Mozifian, J.~Lee, A.~Harakeh, and S.~L. Waslander, ``Joint 3d
  proposal generation and object detection from view aggregation,'' in
  \emph{2018 IEEE/RSJ International Conference on Intelligent Robots and
  Systems (IROS)}.\hskip 1em plus 0.5em minus 0.4em\relax IEEE, 2018, pp. 1--8.

\bibitem{shi2020pv}
S.~Shi, C.~Guo, L.~Jiang, Z.~Wang, J.~Shi, X.~Wang, and H.~Li, ``Pv-rcnn:
  Point-voxel feature set abstraction for 3d object detection,'' in
  \emph{Proceedings of the IEEE conference on computer vision and pattern
  recognition}, 2020, pp. 10\,529--10\,538.

\bibitem{yang20203dssd}
Z.~Yang, Y.~Sun, S.~Liu, and J.~Jia, ``3dssd: Point-based 3d single stage
  object detector,'' in \emph{Proceedings of the IEEE conference on computer
  vision and pattern recognition}, 2020, pp. 11\,040--11\,048.

\bibitem{liu2016ssd}
W.~Liu, D.~Anguelov, D.~Erhan, C.~Szegedy, S.~Reed, C.-Y. Fu, and A.~C. Berg,
  ``Ssd: Single shot multibox detector,'' in \emph{Proceedings of the European
  Conference on Computer Vision (ECCV)}, 2016, pp. 21--37.

\bibitem{lin2017focal}
T.-Y. Lin, P.~Goyal, R.~Girshick, K.~He, and P.~Doll{\'a}r, ``Focal loss for
  dense object detection,'' in \emph{Proceedings of the IEEE international
  conference on computer vision}, 2017, pp. 2980--2988.

\bibitem{ren2015faster}
S.~Ren, K.~He, R.~Girshick, and J.~Sun, ``Faster r-cnn: Towards real-time
  object detection with region proposal networks,'' \emph{Advances in neural
  information processing systems}, vol.~28, pp. 91--99, 2015.

\bibitem{dai2016r}
J.~Dai, Y.~Li, K.~He, and J.~Sun, ``R-fcn: Object detection via region-based
  fully convolutional networks,'' in \emph{Advances in neural information
  processing systems}, 2016, pp. 379--387.

\bibitem{liang2018deep}
M.~Liang, B.~Yang, S.~Wang, and R.~Urtasun, ``Deep continuous fusion for
  multi-sensor 3d object detection,'' in \emph{Proceedings of the European
  Conference on Computer Vision (ECCV)}, 2018, pp. 641--656.

\bibitem{qi2018frustum}
C.~R. Qi, W.~Liu, C.~Wu, H.~Su, and L.~J. Guibas, ``Frustum pointnets for 3d
  object detection from rgb-d data,'' in \emph{Proceedings of the IEEE
  conference on computer vision and pattern recognition}, 2018, pp. 918--927.

\bibitem{zhang2020pc}
Y.~Zhang, D.~Huang, and Y.~Wang, ``Pc-rgnn: Point cloud completion and graph
  neural network for 3d object detection,'' \emph{Proceedings of the AAAI
  Conference on Artificial Intelligence}, 2021.

\bibitem{geiger2012we}
A.~Geiger, P.~Lenz, and R.~Urtasun, ``Are we ready for autonomous driving? the
  kitti vision benchmark suite,'' in \emph{Proceedings of the IEEE conference
  on computer vision and pattern recognition}.\hskip 1em plus 0.5em minus
  0.4em\relax IEEE, 2012, pp. 3354--3361.

\bibitem{sun2020scalability}
P.~Sun, H.~Kretzschmar, X.~Dotiwalla, A.~Chouard, V.~Patnaik, P.~Tsui, J.~Guo,
  Y.~Zhou, Y.~Chai, B.~Caine \emph{et~al.}, ``Scalability in perception for
  autonomous driving: Waymo open dataset,'' in \emph{Proceedings of the IEEE
  conference on computer vision and pattern recognition}, 2020, pp. 2446--2454.

\bibitem{he2016deep}
K.~He, X.~Zhang, S.~Ren, and J.~Sun, ``Deep residual learning for image
  recognition,'' in \emph{Proceedings of the IEEE conference on computer vision
  and pattern recognition}, 2016, pp. 770--778.

\bibitem{qi2017pointnet++}
C.~R. Qi, L.~Yi, H.~Su, and L.~J. Guibas, ``Pointnet++: Deep hierarchical
  feature learning on point sets in a metric space,'' in \emph{Advances in
  neural information processing systems}, 2017, pp. 5099--5108.

\bibitem{mopuri2018cnn}
K.~R. Mopuri, U.~Garg, and R.~V. Babu, ``Cnn fixations: an unraveling approach
  to visualize the discriminative image regions,'' \emph{IEEE Transactions on
  Image Processing}, vol.~28, no.~5, pp. 2116--2125, 2018.

\bibitem{cotnet}
Y.~Li, T.~Yao, Y.~Pan, and T.~Mei, ``Contextual transformer networks for visual
  recognition,'' \emph{arXiv preprint arXiv:2107.12292}, 2021.

\bibitem{cai2020joint}
Q.~Cai, Y.~Wang, Y.~Pan, T.~Yao, and T.~Mei, ``Joint contrastive learning with
  infinite possibilities,'' \emph{Advances in Neural Information Processing
  Systems}, vol.~33, pp. 12\,638--12\,648, 2020.

\bibitem{bao2019monofenet}
W.~Bao, B.~Xu, and Z.~Chen, ``Monofenet: Monocular 3d object detection with
  feature enhancement networks,'' \emph{IEEE Transactions on Image Processing},
  vol.~29, pp. 2753--2765, 2019.

\bibitem{rahman2019notice}
M.~M. Rahman, Y.~Tan, J.~Xue, and K.~Lu, ``Notice of violation of ieee
  publication principles: Recent advances in 3d object detection in the era of
  deep neural networks: A survey,'' \emph{IEEE Transactions on Image
  Processing}, vol.~29, pp. 2947--2962, 2019.

\bibitem{feng2020relation}
M.~Feng, S.~Z. Gilani, Y.~Wang, L.~Zhang, and A.~Mian, ``Relation graph network
  for 3d object detection in point clouds,'' \emph{IEEE Transactions on Image
  Processing}, vol.~30, pp. 92--107, 2020.

\bibitem{deng2020voxel}
J.~Deng, S.~Shi, P.~Li, W.~Zhou, Y.~Zhang, and H.~Li, ``Voxel r-cnn: Towards
  high performance voxel-based 3d object detection,'' \emph{Proceedings of the
  AAAI Conference on Artificial Intelligence}, 2021.

\bibitem{lahoud20172d}
J.~Lahoud and B.~Ghanem, ``2d-driven 3d object detection in rgb-d images,'' in
  \emph{Proceedings of the IEEE international conference on computer vision},
  2017, pp. 4622--4630.

\bibitem{qi2019deep}
C.~R. Qi, O.~Litany, K.~He, and L.~J. Guibas, ``Deep hough voting for 3d object
  detection in point clouds,'' in \emph{Proceedings of the IEEE international
  conference on computer vision}, 2019, pp. 9277--9286.

\bibitem{shi2020points}
S.~Shi, Z.~Wang, J.~Shi, X.~Wang, and H.~Li, ``From points to parts: 3d object
  detection from point cloud with part-aware and part-aggregation network,''
  \emph{IEEE transactions on pattern analysis and machine intelligence}, 2020.

\bibitem{wang2019frustum}
Z.~Wang and K.~Jia, ``Frustum convnet: Sliding frustums to aggregate local
  point-wise features for amodal 3d object detection,'' in \emph{2019 IEEE/RSJ
  International Conference on Intelligent Robots and Systems (IROS)}.\hskip 1em
  plus 0.5em minus 0.4em\relax IEEE, 2019, pp. 1742--1749.

\bibitem{Yang_2019_ICCV}
Z.~Yang, Y.~Sun, S.~Liu, X.~Shen, and J.~Jia, ``Std: Sparse-to-dense 3d object
  detector for point cloud,'' in \emph{Proceedings of the IEEE international
  conference on computer vision}, 2019, pp. 1951--1960.

\bibitem{engelcke2017vote3deep}
M.~Engelcke, D.~Rao, D.~Z. Wang, C.~H. Tong, and I.~Posner, ``Vote3deep: Fast
  object detection in 3d point clouds using efficient convolutional neural
  networks,'' in \emph{2017 IEEE International Conference on Robotics and
  Automation (ICRA)}, 2017, pp. 1355--1361.

\bibitem{He_2020_CVPR}
C.~He, H.~Zeng, J.~Huang, X.-S. Hua, and L.~Zhang, ``Structure aware
  single-stage 3d object detection from point cloud,'' in \emph{Proceedings of
  the IEEE conference on computer vision and pattern recognition}, 2020, pp.
  11\,873--11\,882.

\bibitem{lang2019pointpillars}
A.~H. Lang, S.~Vora, H.~Caesar, L.~Zhou, J.~Yang, and O.~Beijbom,
  ``Pointpillars: Fast encoders for object detection from point clouds,'' in
  \emph{Proceedings of the IEEE conference on computer vision and pattern
  recognition}, 2019, pp. 12\,697--12\,705.

\bibitem{yang2018pixor}
B.~Yang, W.~Luo, and R.~Urtasun, ``Pixor: Real-time 3d object detection from
  point clouds,'' in \emph{Proceedings of the IEEE conference on computer
  vision and pattern recognition}, 2018, pp. 7652--7660.

\bibitem{deng2021multi}
J.~Deng, W.~Zhou, Y.~Zhang, and H.~Li, ``From multi-view to hollow-3d:
  Hallucinated hollow-3d r-cnn for 3d object detection,'' \emph{IEEE
  Transactions on Circuits and Systems for Video Technology}, vol.~31, no.~12,
  pp. 4722--4734, 2021.

\bibitem{shrivastava2016training}
A.~Shrivastava, A.~Gupta, and R.~Girshick, ``Training region-based object
  detectors with online hard example mining,'' in \emph{Proceedings of the IEEE
  conference on computer vision and pattern recognition}, 2016, pp. 761--769.

\bibitem{cao2020prime}
Y.~Cao, K.~Chen, C.~C. Loy, and D.~Lin, ``Prime sample attention in object
  detection,'' in \emph{Proceedings of the IEEE conference on computer vision
  and pattern recognition}, 2020, pp. 11\,583--11\,591.

\bibitem{pang2019libra}
J.~Pang, K.~Chen, J.~Shi, H.~Feng, W.~Ouyang, and D.~Lin, ``Libra r-cnn:
  Towards balanced learning for object detection,'' in \emph{Proceedings of the
  IEEE conference on computer vision and pattern recognition}, 2019, pp.
  821--830.

\bibitem{cai2020learning}
Q.~Cai, Y.~Pan, Y.~Wang, J.~Liu, T.~Yao, and T.~Mei, ``Learning a unified
  sample weighting network for object detection,'' in \emph{Proceedings of the
  IEEE conference on computer vision and pattern recognition}, 2020, pp.
  14\,173--14\,182.

\bibitem{rezatofighi2019generalized}
H.~Rezatofighi, N.~Tsoi, J.~Gwak, A.~Sadeghian, I.~Reid, and S.~Savarese,
  ``Generalized intersection over union: A metric and a loss for bounding box
  regression,'' in \emph{Proceedings of the IEEE/CVF conference on computer
  vision and pattern recognition}, 2019, pp. 658--666.

\bibitem{feng2021tood}
C.~Feng, Y.~Zhong, Y.~Gao, M.~R. Scott, and W.~Huang, ``Tood: Task-aligned
  one-stage object detection,'' in \emph{Proceedings of the IEEE/CVF
  International Conference on Computer Vision}, 2021, pp. 3510--3519.

\bibitem{zhang2021varifocalnet}
H.~Zhang, Y.~Wang, F.~Dayoub, and N.~Sunderhauf, ``Varifocalnet: An iou-aware
  dense object detector,'' in \emph{Proceedings of the IEEE/CVF Conference on
  Computer Vision and Pattern Recognition}, 2021, pp. 8514--8523.

\bibitem{dai2017deformable}
J.~Dai, H.~Qi, Y.~Xiong, Y.~Li, G.~Zhang, H.~Hu, and Y.~Wei, ``Deformable
  convolutional networks,'' in \emph{Proceedings of the IEEE international
  conference on computer vision}, 2017, pp. 764--773.

\bibitem{he2017mask}
K.~He, G.~Gkioxari, P.~Doll{\'a}r, and R.~Girshick, ``Mask r-cnn,'' in
  \emph{Proceedings of the IEEE international conference on computer vision},
  2017, pp. 2961--2969.

\bibitem{cai2019exploring}
Q.~Cai, Y.~Pan, C.-W. Ngo, X.~Tian, L.~Duan, and T.~Yao, ``Exploring object
  relation in mean teacher for cross-domain detection,'' in \emph{Proceedings
  of the IEEE/CVF Conference on Computer Vision and Pattern Recognition}, 2019,
  pp. 11\,457--11\,466.

\bibitem{deng2021minet}
J.~Deng, Y.~Pan, T.~Yao, W.~Zhou, H.~Li, and T.~Mei, ``Minet: Meta-learning
  instance identifiers for video object detection,'' \emph{IEEE Transactions on
  Image Processing}, vol.~30, pp. 6879--6891, 2021.

\bibitem{redmon2016you}
J.~Redmon, S.~Divvala, R.~Girshick, and A.~Farhadi, ``You only look once:
  Unified, real-time object detection,'' in \emph{Proceedings of the IEEE
  conference on computer vision and pattern recognition}, 2016, pp. 779--788.

\bibitem{deng2020single}
J.~Deng, Y.~Pan, T.~Yao, W.~Zhou, H.~Li, and T.~Mei, ``Single shot video object
  detector,'' \emph{IEEE Transactions on Multimedia}, vol.~23, pp. 846--858,
  2020.

\bibitem{cai2018cascade}
Z.~Cai and N.~Vasconcelos, ``Cascade r-cnn: Delving into high quality object
  detection,'' in \emph{Proceedings of the IEEE conference on computer vision
  and pattern recognition}, 2018, pp. 6154--6162.

\bibitem{chen2019hybrid}
K.~Chen, J.~Pang, J.~Wang, Y.~Xiong, X.~Li, S.~Sun, W.~Feng, Z.~Liu, J.~Shi,
  W.~Ouyang \emph{et~al.}, ``Hybrid task cascade for instance segmentation,''
  in \emph{Proceedings of the IEEE conference on computer vision and pattern
  recognition}, 2019, pp. 4974--4983.

\bibitem{gidaris2015object}
S.~Gidaris and N.~Komodakis, ``Object detection via a multi-region and semantic
  segmentation-aware cnn model,'' in \emph{Proceedings of the IEEE
  international conference on computer vision}, 2015, pp. 1134--1142.

\bibitem{gidaris2016attend}
------, ``Attend refine repeat: Active box proposal generation via in-out
  localization,'' in \emph{British Machine Vision Conference}, 2016.

\bibitem{yang2016craft}
B.~Yang, J.~Yan, Z.~Lei, and S.~Z. Li, ``Craft objects from images,'' in
  \emph{Proceedings of the IEEE conference on computer vision and pattern
  recognition}, 2016, pp. 6043--6051.

\bibitem{singh2018r}
B.~Singh, H.~Li, A.~Sharma, and L.~S. Davis, ``R-fcn-3000 at 30fps: Decoupling
  detection and classification,'' in \emph{Proceedings of the IEEE conference
  on computer vision and pattern recognition}, 2018, pp. 1081--1090.

\bibitem{fan2019siamese}
H.~Fan and H.~Ling, ``Siamese cascaded region proposal networks for real-time
  visual tracking,'' in \emph{Proceedings of the IEEE conference on computer
  vision and pattern recognition}, 2019, pp. 7952--7961.

\bibitem{Yang_2020_CVPR}
Z.~Yang, Y.~Sun, S.~Liu, and J.~Jia, ``3dssd: Point-based 3d single stage
  object detector,'' in \emph{Proceedings of the IEEE conference on computer
  vision and pattern recognition}, 2020, pp. 11\,040--11\,048.

\bibitem{liu2020tanet}
Z.~Liu, X.~Zhao, T.~Huang, R.~Hu, Y.~Zhou, and X.~Bai, ``Tanet: Robust 3d
  object detection from point clouds with triple attention.'' in
  \emph{Proceedings of the AAAI Conference on Artificial Intelligence}, 2020,
  pp. 11\,677--11\,684.

\bibitem{Shi_2020_CVPR}
S.~Shi, C.~Guo, L.~Jiang, Z.~Wang, J.~Shi, X.~Wang, and H.~Li, ``Pv-rcnn:
  Point-voxel feature set abstraction for 3d object detection,'' in
  \emph{Proceedings of the IEEE conference on computer vision and pattern
  recognition}, 2020, pp. 10\,529--10\,538.

\bibitem{zheng2020cia}
W.~Zheng, W.~Tang, S.~Chen, L.~Jiang, and C.-W. Fu, ``Cia-ssd: Confident
  iou-aware single-stage object detector from point cloud,'' \emph{Proceedings
  of the AAAI Conference on Artificial Intelligence}, 2021.

\bibitem{kingma2014adam}
D.~P. Kingma and J.~Ba, ``Adam: A method for stochastic optimization,''
  \emph{International Conference on Learning Representations}, 2015.

\bibitem{zhao20193d}
X.~Zhao, Z.~Liu, R.~Hu, and K.~Huang, ``3d object detection using scale
  invariant and feature reweighting networks,'' in \emph{Proceedings of the
  AAAI Conference on Artificial Intelligence}, 2019, pp. 9267--9274.

\bibitem{liang2019multi}
M.~Liang, B.~Yang, Y.~Chen, R.~Hu, and R.~Urtasun, ``Multi-task multi-sensor
  fusion for 3d object detection,'' in \emph{Proceedings of the IEEE conference
  on computer vision and pattern recognition}, 2019, pp. 7345--7353.

\bibitem{Ye_2020_CVPR}
M.~Ye, S.~Xu, and T.~Cao, ``Hvnet: Hybrid voxel network for lidar based 3d
  object detection,'' in \emph{Proceedings of the IEEE conference on computer
  vision and pattern recognition}, 2020, pp. 1631--1640.

\bibitem{zhou2020end}
Y.~Zhou, P.~Sun, Y.~Zhang, D.~Anguelov, J.~Gao, T.~Ouyang, J.~Guo, J.~Ngiam,
  and V.~Vasudevan, ``End-to-end multi-view fusion for 3d object detection in
  lidar point clouds,'' in \emph{Conference on Robot Learning}, 2019, pp.
  923--932.

\bibitem{wang2020pillar}
Y.~Wang, A.~Fathi, A.~Kundu, D.~Ross, C.~Pantofaru, T.~Funkhouser, and
  J.~Solomon, ``Pillar-based object detection for autonomous driving,''
  \emph{Proceedings of the European Conference on Computer Vision (ECCV)},
  2020.

\bibitem{ge2020afdet}
R.~Ge, Z.~Ding, Y.~Hu, Y.~Wang, S.~Chen, L.~Huang, and Y.~Li, ``Afdet: Anchor
  free one stage 3d object detection,'' \emph{IEEE Conference on Computer
  Vision and Pattern Recognition Workshops (CVPRW)}, 2020.

\end{thebibliography}
	}

\end{document}